\documentclass[letterpaper]{article} 
\usepackage{aaai2026}  
\usepackage{times}  
\usepackage{helvet}  
\usepackage{courier}  
\usepackage[hyphens]{url}  
\usepackage{graphicx} 
\usepackage{natbib}  
\usepackage{caption} 
\usepackage{algorithm}
\usepackage{algorithmic}

\usepackage{newfloat}
\usepackage{listings}
\DeclareCaptionStyle{ruled}{labelfont=normalfont,labelsep=colon,strut=off} 
\floatstyle{ruled}
\newfloat{listing}{tb}{lst}{}
\floatname{listing}{Listing}
\nocopyright

\usepackage{makecell}
\usepackage{amssymb}
\usepackage{pifont} 
\usepackage{booktabs}
\usepackage{xspace}
\usepackage{subcaption}
\usepackage{multirow}
\usepackage{amsmath}
\usepackage{dsfont}

\newcommand{\real}{\mathds{R}}

\makeatletter
\DeclareRobustCommand\onedot{\futurelet\@let@token\@onedot}
\def\@onedot{\ifx\@let@token.\else.\null\fi\xspace}
\def\eg{\emph{e.g}\onedot}

\makeatother

\newcommand{\ourdatasetname}{\textit{MultiObjectBlender}\xspace}

\def \customparskip {0.2em}
\newcommand{\customparagraph}[1]{\vspace{\customparskip}\textbf{#1}}

\usepackage{enumitem}

\usepackage[capitalise]{cleveref}
\crefname{section}{Sec.}{Secs.}

\title{ROODI: \underline{R}econstructing \underline{O}ccluded \underline{O}bjects with \underline{D}enoising \underline{I}npainters}

\author{
Yeonjin Chang\textsuperscript{\rm 1},
Erqun Dong\textsuperscript{\rm 2}, 
Seunghyeon Seo\textsuperscript{\rm 1},
Nojun Kwak\textsuperscript{\rm 1},
Kwang Moo Yi\textsuperscript{\rm 2} \\
}
\affiliations {
    \textsuperscript{\rm 1}Seoul National University,
    \textsuperscript{\rm 2}University of British Columbia
}

\begin{document}
\maketitle

\begin{abstract}
While the quality of novel-view images has improved dramatically with 3D Gaussian Splatting, extracting specific objects from scenes remains challenging.
Isolating individual 3D Gaussian primitives for each object and handling occlusions in scenes remains far from being solved.
We propose a novel object extraction method based on two key principles: (1) object-centric reconstruction through removal of irrelevant primitives; and (2) leveraging generative inpainting to compensate for missing observations caused by occlusions.
For pruning, we propose to remove irrelevant Gaussians by looking into how close they are to its K-nearest neighbors and removing those that are statistical outliers.
Importantly, these distances must take into account the actual spatial extent they cover---we thus propose to use Wasserstein distances.
For inpainting, we employ an off-the-shelf diffusion-based inpainter combined with occlusion reasoning, utilizing the 3D representation of the entire scene. 
Our findings highlight the crucial synergy between proper pruning and inpainting, both of which significantly enhance extraction performance.
We evaluate our method on a standard real-world dataset and introduce a synthetic dataset for quantitative analysis. 
Our approach outperforms the state-of-the-art, demonstrating its effectiveness in object extraction from complex scenes.
\end{abstract}

\section{Introduction}
\label{sec:intro}
\begin{figure}[t]
    \centering
    \newcommand{\imgw}{0.88\linewidth}
        \includegraphics[width=\imgw]{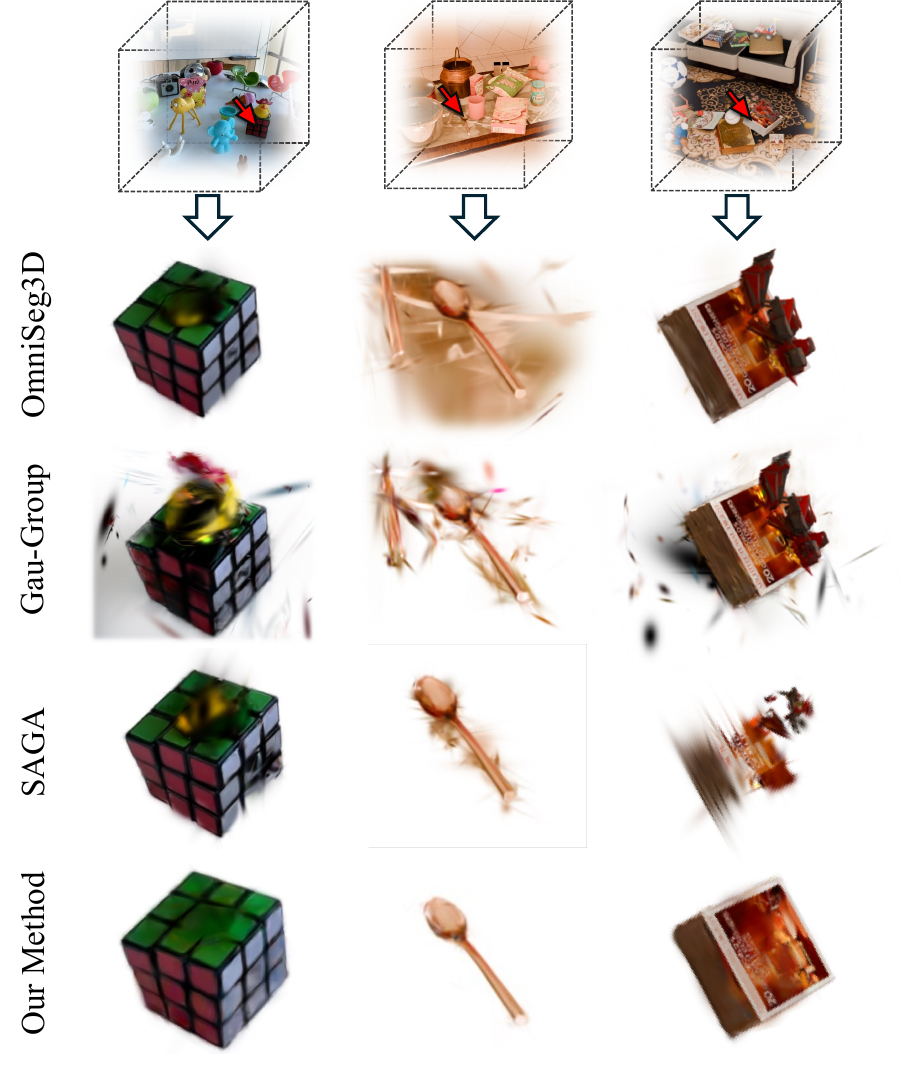}
    \caption{\textbf{Qualitative highlight -- }
        We propose a method to reconstruct specific objects in complex scenes from posed images, even if they are occluded.
        To reduce floaters, we prune Gaussians that are not of the object of interest, and use an off-the-shelf denoising diffusion inpainter together with occlusion reasoning.
        Our method provides significantly improved results compared to the state of the art.
    }
    \label{fig:teaser}
\end{figure}

3D scenes are typically composed of numerous objects.
Naturally, beyond the original 3D Gaussian Splatting (3DGS)~\cite{3dgs}, several works attempt to interpret 3D scenes at the object level to better capture their semantic structure~\cite{gaussiangrouping, clickgaussian, saga, lifting, feature3dgs, OmniSeg3D}.
These studies primarily train semantic features of 3D scenes by leveraging semantic maps from 2D segmentation models, modeling the overall structure and semantics of the scene.
However, extracting the 3DGS model of target objects with these semantic features introduces additional challenges---precise separation of the target object from the rest of the scene is difficult and (self-)occlusions further make this problem harder.
As shown in \cref{fig:teaser}, none of the existing methods yield a clearly extracted target model and have so-called \textit{floaters}.

In this work, we propose a novel method for 3D object extraction, based on three key ideas:
(1) regardless of the choice of which object, scene geometry must remain the same---the full scene geometry serves as base knowledge for object extraction,
(2) objects are \textit{local}---we are interested in a single cluster of primitives that form a single object, and 
(3) occlusions can easily be reasoned by comparing the primitives of the target object with those of the full scene.
Based on (1), we adopt a two-stage pipeline that trains the scene geometry first, followed by additional training of object features to distinguish objects. 
Based on (2), we look into the local structure of the primitives and prune those that are not part of the target object.
We propose to use the distance of each Gaussian to their nearest neighbors and remove those that are statistical outliers.
As Gaussians are entities that cover a \emph{region} in 3D space, we consider the distances in terms of Wasserstein distances---we show later in \cref{sec:ablation} that this is important for proper pruning.
From (3), we use a pretrained diffusion model to inpaint the occluded regions of the target object.
This leads to clear object extraction, one that is robust to occlusions and floaters; see \cref{fig:teaser}.

In more detail, we first train a 3DGS model of the scene, then extract the 3DGS model of the target objects in the second stage. 
To extract an object, we start with Segment Anything Model 2 (SAM2)~\cite{sam, ravi2024sam} to obtain the segmentation map of the object of interest.
We then distill this map into the 3DGS model by briefly fine-tuning the model for 500 iterations.
We then update 3D Gaussians specific to the target object by:
(1) pruning non-object Gaussians by removing those that are statistical outliers when looking into their Wasserstein distances to nearest neighbors;
(2) rendering both the target object and the full scene to identify occluded regions of the target object; and
(3) inpainting the occluded regions of the target object with a pretrained inpainting diffusion model~\cite{podell2023sdxl, diffusers-inpainting}; and finally
(4) updating the parameters of the 3D Gaussians to better match the inpainted target object.
Notably, all of these steps are crucial, as without pruning floaters, the diffusion model is unable to recognize or inpaint the target object due to floaters, and inpainting is further critical to obtain clear, sharp images---we show empirically that simple general diffusion-based enhancement is insufficient.

To evaluate our method, we conduct experiments on both synthetic and real-world datasets.
We use the standard LERF dataset~\cite{lerf} to allow qualitative comparison with existing methods.
As the real-world dataset does not have a ground-truth object-centric extraction (e.g., parts of object can be occluded), we create a new synthetic dataset, named \ourdatasetname.
We render images from a 3D scene composed of many objects from ShapeNet~\cite{shapenet} using Blender~\cite{blender}, where many objects are occluding each other.
We then render both the full scene and only the target object, the latter of which can be used to quantitatively evaluate the extraction quality.
Our method significantly outperforms existing methods on both datasets.

\customparagraph{Contributions.}
To summarize, in this work:
\begin{itemize}[itemsep=0pt, topsep=0pt]
    \item we propose a method to extract objects within a 3DGS model that prunes non-object Gaussians through statistical outlier removal, and performs occlusion reasoning and inpainting;
    \item we propose to use Wasserstein distances when measuring distances between Gaussians for this purpose;
    \item we demonstrate that the two-stage pipeline, the removal of non-object Gaussians, and the inpainting of occluded regions are all essential;
    \item we create a new synthetic dataset to evaluate the performance of object extraction methods;\footnote{We will make this dataset publicly available.}
    \item we show that our method significantly outperforms existing methods on both synthetic and real-world datasets.
\end{itemize}

\section{Related Work}
\customparagraph{3D reconstruction.}
Research on 3D scene reconstruction from multi-view images has gained momentum since NeRF~\cite{nerf}, which introduced an implicit representation of 3D space using neural networks, enabling photo-realistic novel view synthesis.
However, NeRF suffers from long training and inference times, which led to subsequent research incorporating voxel grids~\cite{yu2021plenoctrees, fridovich2022plenoxels, hu2022efficientnerf, sun2022direct}, tri-planes~\cite{khatib2024trinerflet, fridovich2023k, chen2022tensorf}, and hybrid representations~\cite{turki2024hybridnerf, zhang2025efficient}.

In contrast to the implicit formulation of NeRFs, 3D Gaussian Splatting (3DGS)~\cite{3dgs} emerged as an explicit representation that models the 3D radiance field using Gaussian primitives.
This approach enables real-time rendering while achieving high visual quality.
Following the introduction of 3DGS, research has rapidly expanded across various topics and applications, including dynamic scenes with moving objects~\cite{luiten2024dynamic, li2024spacetime, yang2023real, yang2024deformable, wu20244d}, 3D/4D object generation~\cite{tang2023dreamgaussian, yi2024gaussiandreamer, chen2024text, ren2023dreamgaussian4d, zeng2024stag4d, wu2024sc4d, gao2024gaussianflow}, and semantic encoding in representations~\cite{saga, gaussiangrouping, ji2024segment, chen2024gaussianeditor}.
Among these, extracting specific objects within a 3D scene using semantic encoding has become an important topic for scene editing.
Existing methods, however, often struggle with floating artifacts or occlusions as shown in \cref{fig:teaser}, making it challenging to achieve clean and accurate extractions.
In this paper, we address the problem of extracting objects from 3D space while preserving high-quality details.

\begin{figure*}[t]
    \centering
    \includegraphics[width=0.95\textwidth]{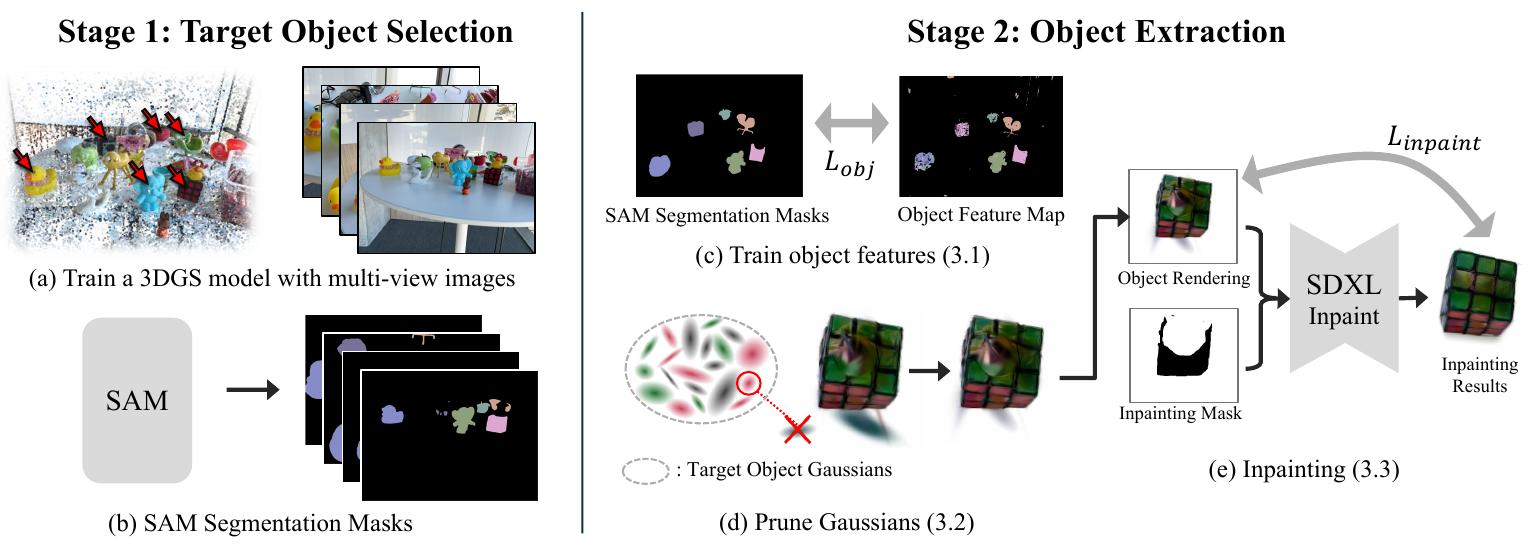}
    \caption{
        {\bf Overview -- }
        (a) Given a standard 3DGS model along with multi-view images, we select the object to extract. 
        (b) We obtain segmentation maps for the target objects using SAM2, (c) to distill the semantic map into the object feature of 3D Gaussians.
        (d) Using the object features, we initially separate the object and prune floaters, via our statistical pruning with Wasserstein distance.
        (e) We then utilize an inpainting diffusion model to fill in the missing parts, reasoning occlusions based on the depth map.
    }
    \label{fig:overview}
\end{figure*}

\customparagraph{3D semantic segmentation.}
We first briefly review semantic segmentation using 3DGS, as many object extraction methods rely on it.
Segmentation on 3DGS is typically achieved by incorporating an additional semantic feature field~\cite{clickgaussian, feature3dgs, gaussiangrouping, saga, contrastivegaussian}, or assigning 3D labels~\cite{flashsplat,lifting}.
These semantic features are trained using 2D semantic maps obtained from foundation models~\cite{sam,lseg} for 2D segmentation. 
Some distill 2D semantic features to 3D semantic features through contrastive learning~\cite{clickgaussian, saga, OmniSeg3D, contrastivegaussian}, while others directly train 3D features using a multichannel renderer~\cite{feature3dgs}.
Alternatively, other methods~\cite{lifting, flashsplat} map 2D semantic masks to 3D labels, and \citet{gaussiangrouping} directly trains semantic features by rendering them and performing pixel-wise classification under the supervision of 2D semantic maps. 
These methods achieve strong semantic segmentation performance when evaluating their rendered 2D semantic maps against those obtained from 2D foundation models. 
However, they place less emphasis on segmentation in 3D space itself, which is directly related to extraction quality.

\customparagraph{3D object extraction.}
Many 3D segmentation studies present object extraction as a downstream application, and not as a primary research focus~\cite{gaussiangrouping, clickgaussian, lifting, OmniSeg3D, garfield, saga, contrastivegaussian}.
Since these methods include semantic features, they can extract objects either by using a classifier to distinguish the groups to which primitives belong~\cite{gaussiangrouping} or by applying thresholding based on feature similarity to extract primitives with high similarity~\cite{saga, OmniSeg3D}.
However, most approaches do not output an actual extracted model but select primitives with similar features, often leaving the threshold for the similarity undefined.
This can lead to suboptimal extraction outcomes when used as an object extractor, and can perform poorly when the scene is complex with occlusions.
In this paper, we prioritize achieving high-quality extracted models with inpainted occluded regions.

\section{Method}
\label{sec:method}

We provide an overview of our method in \cref{fig:overview}.
To start with, we pretrain a 3DGS model of the entire scene.
Then, for the objects of interest, single or multiple, we use Segment Anything V2 (SAM2)~\cite{ravi2024sam} with just \emph{one to two} clicks to obtain their segmentation maps from multiple views.
With these masks, we find the subset of Gaussians that correspond to the target object(s) by distilling these masks into the object features of pretrained 3D Gaussians. 
We then look at the distilled object features to obtain object-specific 3D Gaussians.
These Gaussians, however, are crude and contain many artifacts and holes as shown in \cref{fig:overview}d.
The core of our contribution is how we further train these Gaussians to properly represent objects as shown in \cref{fig:overview}e.

This section is organized as follows: we first discuss the initial training process of 3DGS and how we distill masks into the Gaussians in \cref{sec:init};
we then detail our core contributions, object-centric pruning (\cref{sec:pruning}) and how we utilize pre-trained inpainting models with occlusion reasoning (\cref{sec:inpainting}).

\subsection{Initial training}
\label{sec:init}

We begin by following the standard training process of 3D Gaussian Splatting (3DGS)~\cite{3dgs} to construct a 3DGS model of the entire scene, with the more-recent Markov Chain Monte Carlos-based solver~\cite{kheradmand20253d}.
We parameterize each Gaussian by a rotation matrix $R\in\real^{3\times3}$, scaling $S\in\real$, position $X\in\real^3$, opacity $\alpha\in\real$, and color feature $F_c\in\real^{h}$, where $h$ is the number of spherical harmonic coefficients.
We train using a set of images $\mathcal{I} = \{I^1, I^2, ..., I^t\}$, where $I^*\in\real^{H\times W}$. 
To render a pixel $p\in\real^2$, we first sort Gaussians by their depth relative to the camera (rendering view), and then composite using 
$\alpha$-blending. 
We compute the color of pixel $p$ as:
\begin{equation}
  \mathbf{C}(p) = \sum_{i}^{N} c_i \tilde{\alpha}_i \prod_{j=1}^{i-1} (1 - \tilde{\alpha}_j),
\label{eq:blending}
\end{equation}
where $N\in\real$ is the number of Gaussians contributing to the pixel’s color, $c_i\in\real^3$ and $\tilde{\alpha}_i\in\real$ represent the color and the opacity of the $i$-th Gaussian at pixel $p$ after being projected into the rendering view.

The parameters of the Gaussians are updated to minimize both the $\mathcal{L}_1$ loss and the D-SSIM loss~\cite{3dgs},
where the former is a per-pixel loss enforcing colors to be the same, and the latter focuses on the local structure:
\begin{equation}
  \mathcal{L} = (1 - \lambda_\text{D-SSIM})\mathcal{L}_1 + \lambda_\text{D-SSIM} \mathcal{L}_\text{D-SSIM},
\label{eq:loss}
\end{equation}
where $\lambda_\text{D-SSIM}$ is a hyperparameter balancing the two losses which is typically set as 0.2.

We then, as in Gaussian Grouping~\cite{gaussiangrouping} and many others that incorporate semantics to Gaussians, distill the masks as an additional feature to each Gaussian.
Specifically, for each Gaussian, we add an object feature $F_o\in\real^d$ that encodes the object information of the $C$ objects, where $d$ denotes the dimension of the object features.
Then, after rendering the features of the Gaussians, we pass these features through a shallow classifier to generate the object feature map $\hat{O}\in\real^{H\times W\times C}$ of size $H \times W$,  as shown in \cref{fig:overview}c.
We then compare this feature map with the segmentation masks obtained from SAM2~\cite{ravi2024sam}, and minimize the cross entropy between the object feature map $\hat{O}$ and the segmentation masks $O\in\real^{H\times W}$ that contain which class each pixel belongs to:
\begin{equation}
  \mathcal{L}_\text{obj} = -\frac{1}{HW} \sum_{i=1}^{H} \sum_{j=1}^{W} \sum_{c=1}^{C} \mathds{1}(O_{i,j} = c) \log \hat{O}_{c, i, j} ,
\label{eq:obj_loss}
\end{equation}
where $\mathds{1}$ is the indicator function.
After 500 iterations of training, the object feature becomes sufficiently trained, making it possible to reasonably determine which Gaussians belong to which objects.

Now, with the pretrained 3DGS model of both the scene and the target object, we train the object-specific Gaussians further with object-centric pruning and inpainting.

\subsection{Object-centric pruning}
\label{sec:pruning}

A key observation in reconstructing a specific object is that the task itself is about a single object.
However, directly distilling masks or semantics often leads to the distillation being spread across multiple Gaussians.
This is because the Gaussians are typically non-opaque, meaning multiple Gaussians must work together to form surfaces.
This, in turn, results in the appearance of floaters that are irrelevant to the object of interest, as shown earlier in \cref{fig:teaser}.
Given that object-specific 3D Gaussians should represent only the target object, we can effectively `prune' these floaters by simply removing those that do not follow the typical connectivity with nearby Gaussians, that is, statistical outliers.

\customparagraph{Statistical outlier removal.}
To do so, we propose a pruning strategy based on statistical outlier removal~\cite{sor}.
Our core idea is to look into the average distance that a Gaussian forms with its K nearest neighbors, and remove those that do not follow the general trend---those that are in the long tails of the distance distribution.
Specifically, denoting the distance between the two Gaussians as $\Delta(\cdot,\cdot)$, we prune a Gaussian $\mathcal{G}_i$ if it satisfies:
\begin{equation}
    \frac{1}{K} \sum_{j \in \mathcal{N}_i} 
        \Delta\left(\mathcal{G}_i , \mathcal{G}_j\right) 
        > \bar{\mu} + \alpha \cdot \bar{\sigma},
    \label{eq:prune}
\end{equation}
where $\mathcal{N}_i$ are the K nearest neighbors, $\bar{\mu}=\frac{1}{NK}\sum_{i} \sum_{j \in \mathcal{N}_i}\Delta\left(\mathcal{G}_i , \mathcal{G}_j\right)$ is the average distance and $\bar{\sigma}$ is the standard deviation of distances, and $\alpha$ is a hyperparamter controlling the pruning aggressiveness.

\customparagraph{Measuring distance between Gaussians.}
For \cref{eq:prune} to work properly, it is essential that $\Delta(\cdot,\cdot)$ is defined correctly. 
A naive implementation would consider only the Gaussian centers, but this quickly leads to poor performance as Gaussians are often elongated to represent edges.
We thus propose to use squared 2-Wasserstein distance, $W_2^2$.
Denoting Gaussians as $\mathcal{G} = \mathcal{N}(\mu, \Sigma)$, we thus write:
\begin{equation}
\begin{aligned}
&\Delta\left(\mathcal{G}_i , \mathcal{G}_j\right) 
= W_2^2\!\left(\mathcal{G}_i , \mathcal{G}_j\right) \\
&\;= \lVert \mu_i - \mu_j \rVert_2^{\,2} 
+ \mathrm{Tr}\!\Bigl(
        \Sigma_i + \Sigma_j
        - 2\bigl(\Sigma_j^{1/2}\,\Sigma_i\,\Sigma_j^{1/2}\bigr)^{1/2}
    \Bigr),
\end{aligned}
\end{equation}
where $\Sigma_j^{1/2}$ denotes the principal matrix square root of $\Sigma_j$; we use the squared form for efficiency.

\customparagraph{Implementation.}
Our CUDA implementation of the pruning process runs within 5 seconds for each scene, which we run once at 500 iterations---the cost is virtually none compared to the time it takes to train Gaussians.
This results in an average of 1\% of the Gaussians being pruned.

\subsection{Training with occlusion-reasoned inpainting}
\label{sec:inpainting}

When extracting objects from a scene, not all parts of the objects are visible---some parts may be occluded by other objects in the scene, or at the very least, they would be touching the ground.
Interestingly, while occlusion exists, when reconstructing full scenes, as a byproduct, occluded areas can be easily identified.
We thus find these occluded areas in our object-specific model and inpaint them with an off-the-shelf inpainting model.

\begin{figure}
    \centering
    \footnotesize
    \newcommand{\imgw}{0.25\linewidth}
    \resizebox{\linewidth}{!}{
    \begin{tabular}{cccc}
        & \textbf{Rendering} & \textbf{Inpaint mask} & \textbf{Inpainted} \\

        \rotatebox{90}{w/o smoothing} &
         &
        \includegraphics[width=\imgw]{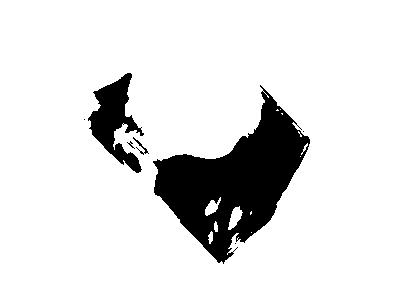} &
        \includegraphics[width=\imgw]{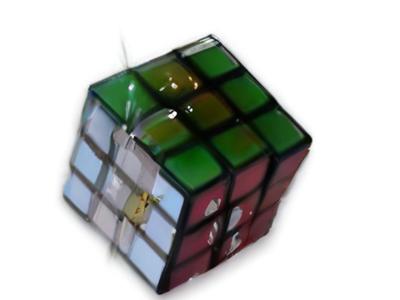} \\

        \rotatebox{90}{Hand-drawn} &
        \includegraphics[width=\imgw]{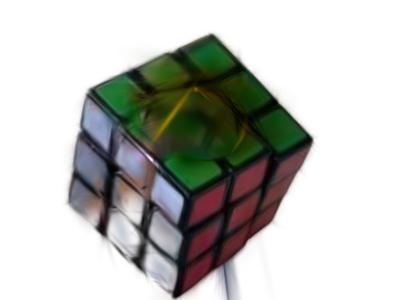} &
        \includegraphics[width=\imgw]{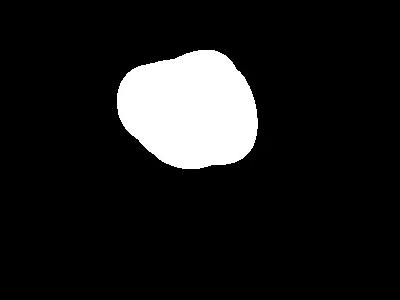} &
        \includegraphics[width=\imgw]{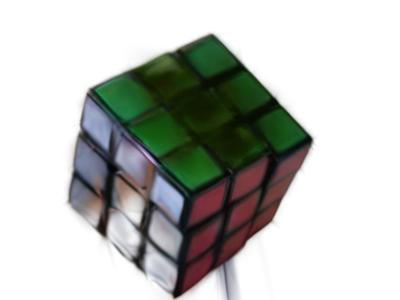} \\  
                
        \rotatebox{90}{Our Method} & 
        &
        \includegraphics[width=\imgw]{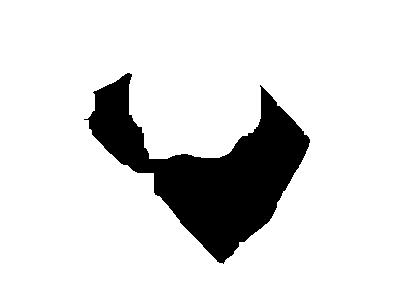} &
        \includegraphics[width=\imgw]{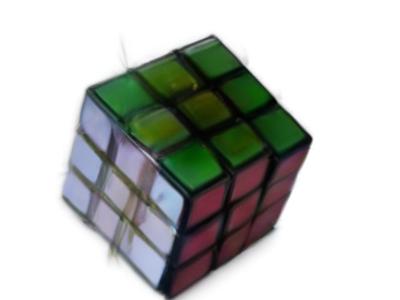} \\
        
    \end{tabular}
    }
    \caption{\textbf{Inpainting mask matters -- }
    The performance of the inpainter can degrade significantly when masks are non-smooth as shown on top, without erosion.
    Even when the mask is precisely designated to the top of the cube, which is not well rendered due to an object on top of the cube (see \cref{fig:qualitative}), the inpainting result can still be suboptimal as rendering can contain floaters.
    Our method smoothes out the inpainting mask with morphological opening, and further allows inpainter to correct object borders by also including the background in the inpainting region.
    }
    \label{fig:mask}
\end{figure}

\customparagraph{Occlusion reasoning.}
Specifically, to identify occluded areas, we look into the depth maps, one rendered using the object-centric 3D Gaussians and the other using the full scene 3D Gaussians.
We then compare the two depth maps and mark pixels where the full scene is closer to the camera than the target object as to-be-inpainted pixels.
We therefore define the mask $\mathbf m$ denoting occluded areas as:
\begin{equation}
  \mathbf m = \mathds{1}(\mathbf d_{\text{scene}} < \mathbf d_{\text{object}}),
\label{eq:mask}
\end{equation}
where $\mathds{1}$ again is the indicator function applied elementwise, and $\mathbf{d}_{\text{scene}}\in\real^{H\times W}$ and $\mathbf{d}_{\text{object}}\in\real^{H\times W}$ are the rendered depth maps of the full scene and the target object, respectively.
As this mask can sometimes be unreliable due to reconstruction errors, we relax the mask to account for errors via morphological opening as shown in \cref{fig:mask} with a $17\times 17$ kernel.
Note that the mask, shown as solid white, also contains the background.
This allows the inpainter to also clean up the borders of the object.

\customparagraph{Inpainting.}
To inpaint occluded areas, we generate four random camera views for each object, with the object centered.
To maintain multi-view consistency, following NeRFiller~\cite{weber2024nerfiller}, we arrange these four renderings into a $2 \times 2$ grid view and provide them to the inpainter simultaneously.
With the occlusion masks for each view from \cref{eq:mask}, we use an off-the-shelf inpainting diffusion model, ``SD-XL Inpainting 0.1'' ~\shortcite{diffusers-inpainting}, which is modified and finetuned from SDXL~\cite{podell2023sdxl}.
We do not use any prompts for the inpainter.
The inpainting outcome, with our mask, provides enhancements as shown in \cref{fig:mask}.
We aggregate these enhancements into the object-centric Gaussians via training, which we detail next.

\customparagraph{Training with inpainted images.}
We then further train the object-centric 3DGS with the resulting four inpainted views.
Similar to the original 3DGS training, we rely on the $\mathcal{L}_1$ loss to guide each pixel.
However, we do not use the D-SSIM loss as we find it to be harmful as inpainting results are not always perfect.
Instead, we use a perceptual loss to guide the inpainting process as in ReconFusion~\cite{reconfusion}: 
\begin{equation}
  \mathcal{L}_\text{inpaint} = \lambda_{L_1-\text{inpaint}}\mathcal{L}_{1-\text{inpaint}} + \lambda_\text{Perc} \mathcal{L}_\text{Perceptual}
  ,
\end{equation}
where $\lambda_{L_1-Inpaint}$ and $\lambda_{Perc}$ are hyperparameters balancing the two losses, and $\mathcal{L}_{Perceptual}$ is the perceptual distance~\cite{zhang2018unreasonable}.
We finetune the object-centric 3DGS for a total of 100 iterations, using the original image (the full scene) every iteration and the inpainted images every other iteration, to prevent inpainting from causing our reconstructions to drift away too much from the original object.

\section{Experiments}

\begin{figure}[t]
    \centering
    \newcommand{\imgw}{0.48\linewidth}
    \begin{subfigure}[b]{\imgw}
        \centering
        \includegraphics[width=\textwidth]{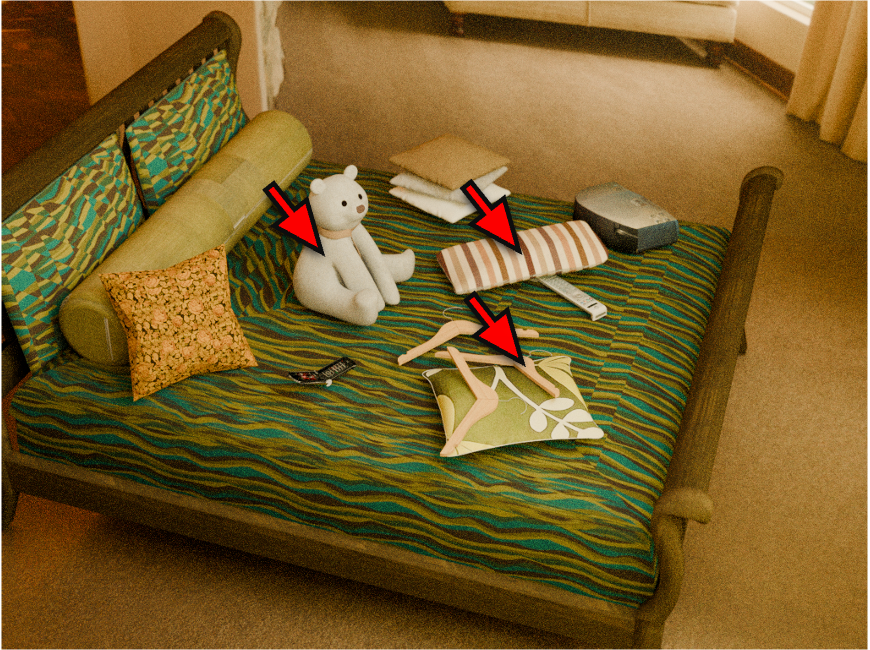}
        \caption{Bedroom}
        \label{fig:item_a}
    \end{subfigure}
    \hfill
    \begin{subfigure}[b]{\imgw}
        \centering
        \includegraphics[width=\textwidth]{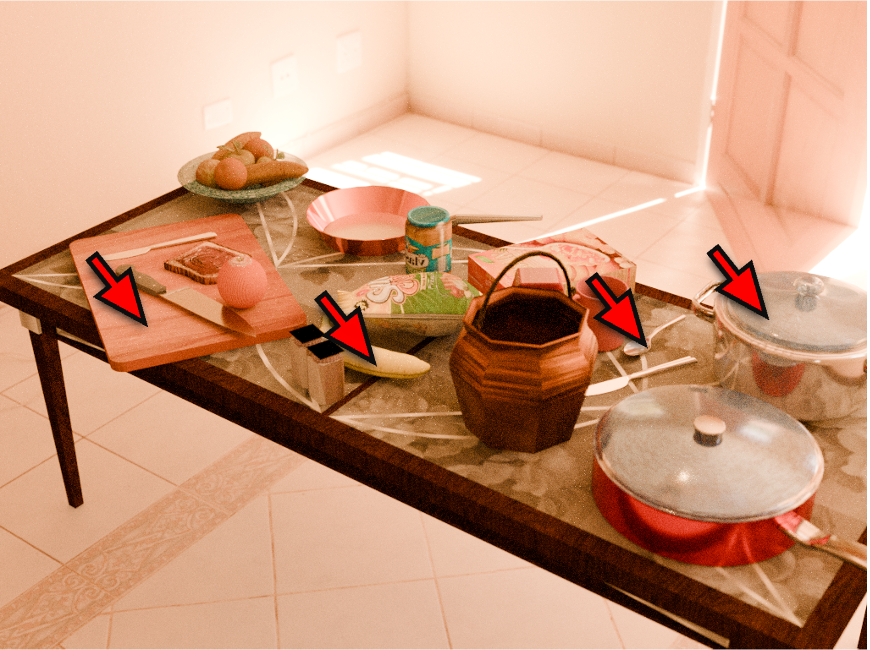}
        \caption{Kitchen}
        \label{fig:item_b}
    \end{subfigure}
    \hfill
    \begin{subfigure}[b]{\imgw}
        \centering
        \includegraphics[width=\textwidth]{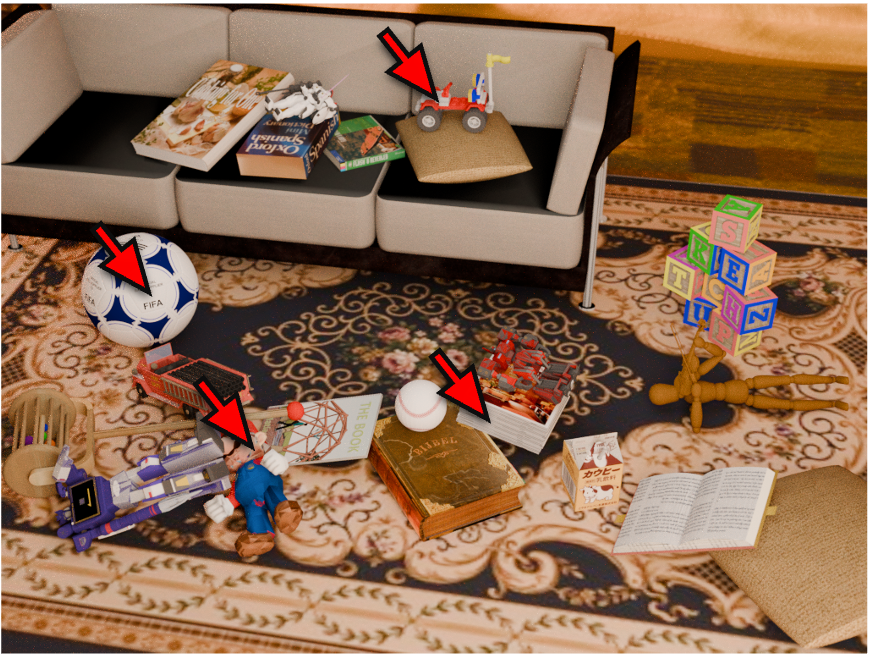}
        \caption{Livingroom}
        \label{fig:item_c}
    \end{subfigure}
    \hfill
    \begin{subfigure}[b]{\imgw}
        \centering
        \includegraphics[width=\textwidth]{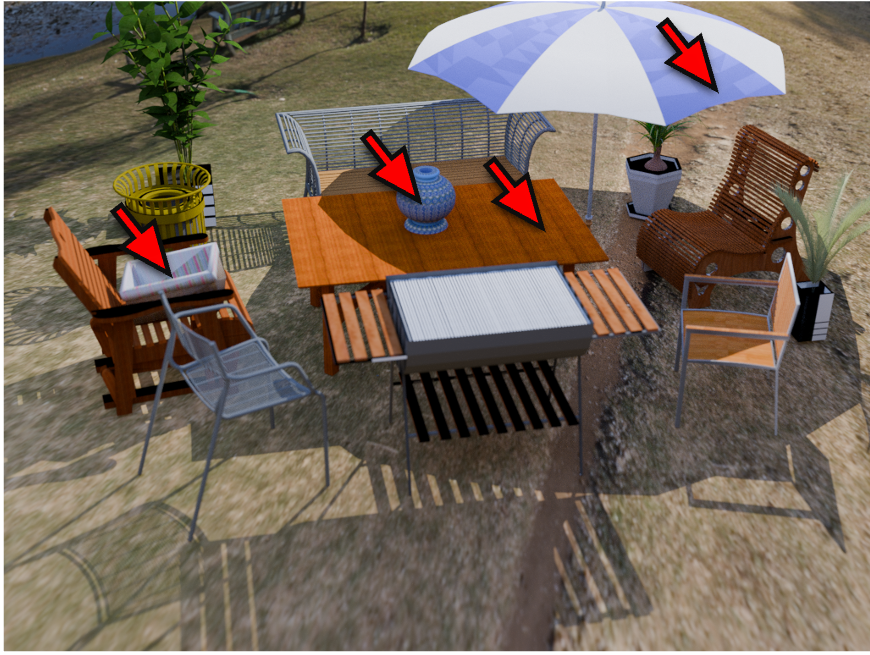}
        \caption{Picnic}
        \label{fig:item_d}
    \end{subfigure}
    \caption{
        \textbf{Example views from \ourdatasetname~-- }
        We synthesize four scenes with ShapeNet~\cite{shapenet} assets.
        In each scene we select three to four objects, each marked with red arrow, as the objects of interest.
        We provide stand-alone renders for each objects as test images.
        }
    \label{fig:multiobjectblender}
\end{figure}

\begin{figure*}[t]
    \centering
    \footnotesize
    \newcommand{\imgw}{0.15\textwidth}
    \setlength{\tabcolsep}{2pt}
    \resizebox{\linewidth}{!}{
    \begin{tabular}{cc ccc cc}
        & \multicolumn{1}{c}{\textbf{Original Image}} & \multicolumn{1}{c}{\textbf{Original (zoomed)}} & \multicolumn{1}{c}{\textbf{Gaussian Grouping}} & \multicolumn{1}{c}{\textbf{SAGA}} & \multicolumn{1}{c}{\textbf{OmniSeg3D-GS}} & \multicolumn{1}{c}{\textbf{Our Method}} \\
        \cmidrule(lr){2-2} \cmidrule(lr){3-3} \cmidrule(lr){4-4} \cmidrule(lr){5-5} \cmidrule(lr){6-6} \cmidrule(lr){7-7}
        
        \raisebox{15pt}{\rotatebox{90}{\makecell{figurines}}} &
        \includegraphics[width=\imgw]{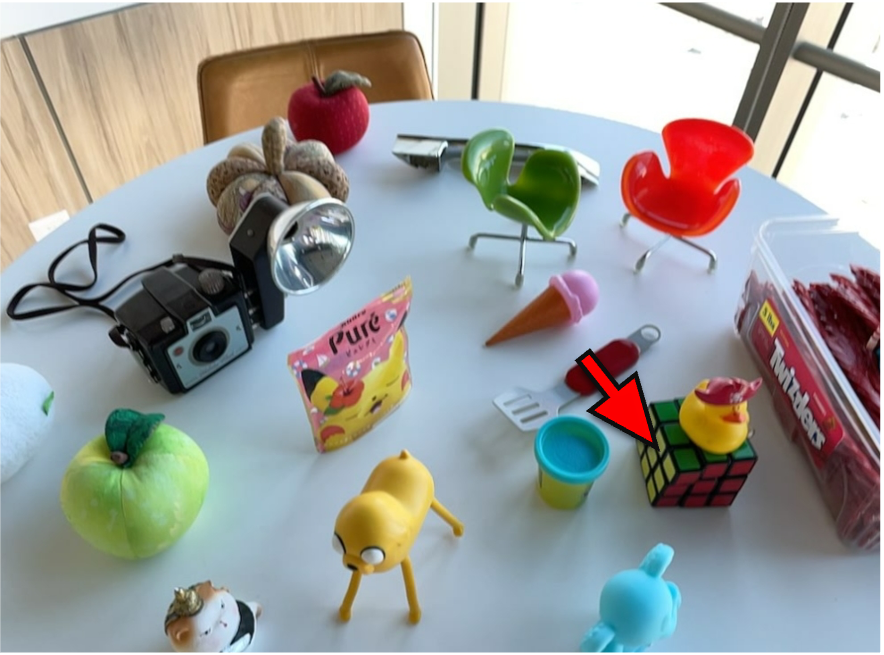} &
        \includegraphics[width=\imgw]{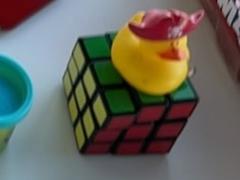} &
        \includegraphics[width=\imgw]{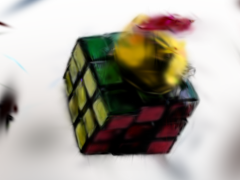} &
        \includegraphics[width=\imgw]{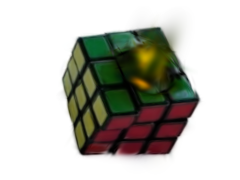} &
        \includegraphics[width=\imgw]{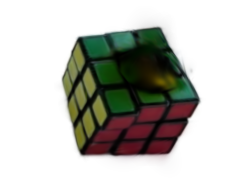} &
        \includegraphics[width=\imgw]{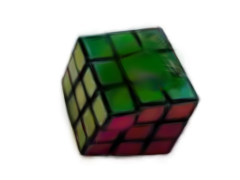} \\
        
        \raisebox{15pt}{\rotatebox{90}{\makecell{ramen}}} &
        \includegraphics[width=\imgw]{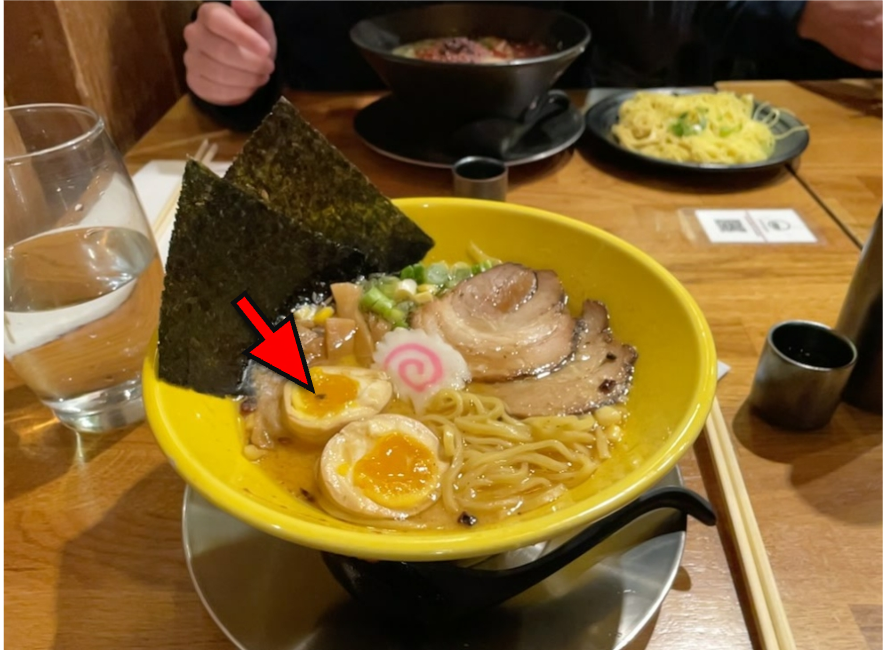} &
        \includegraphics[width=\imgw]{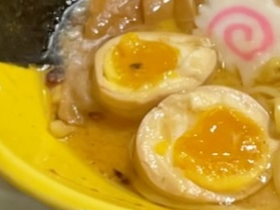} &
        \includegraphics[width=\imgw]{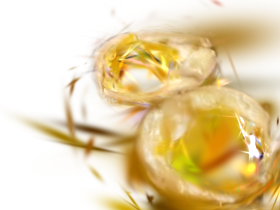} &
        \includegraphics[width=\imgw]{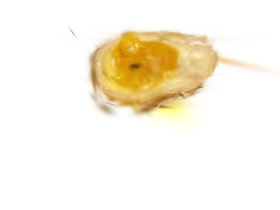} &
        \includegraphics[width=\imgw]{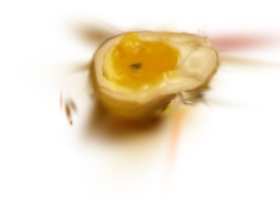} &
        \includegraphics[width=\imgw]{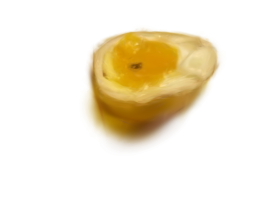} \\
        
        \raisebox{15pt}{\rotatebox{90}{\makecell{teatime}}} &  
        \includegraphics[width=\imgw]{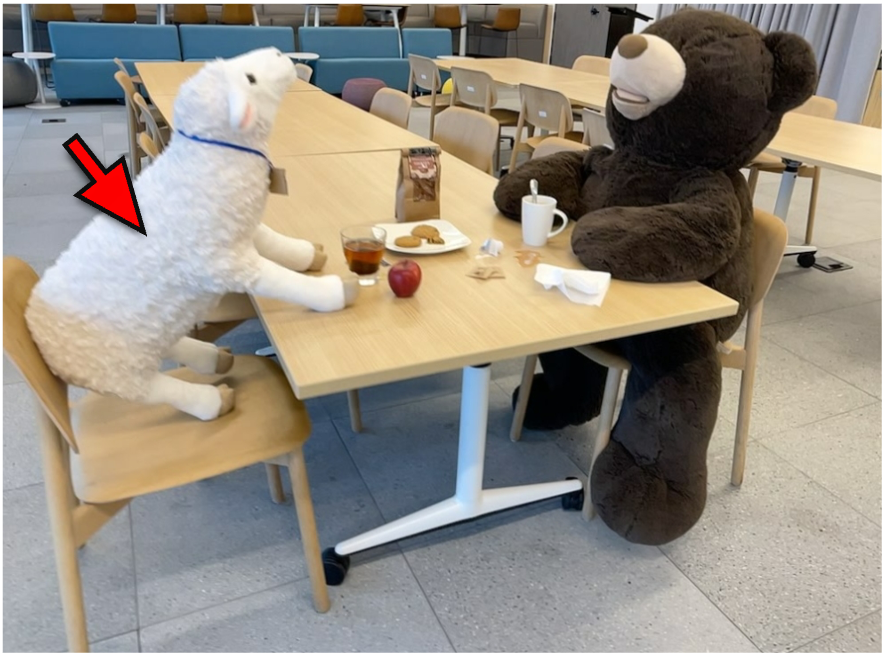} &
        \raisebox{25pt}{\centering \textbf{N/A}} &
        \includegraphics[width=\imgw]{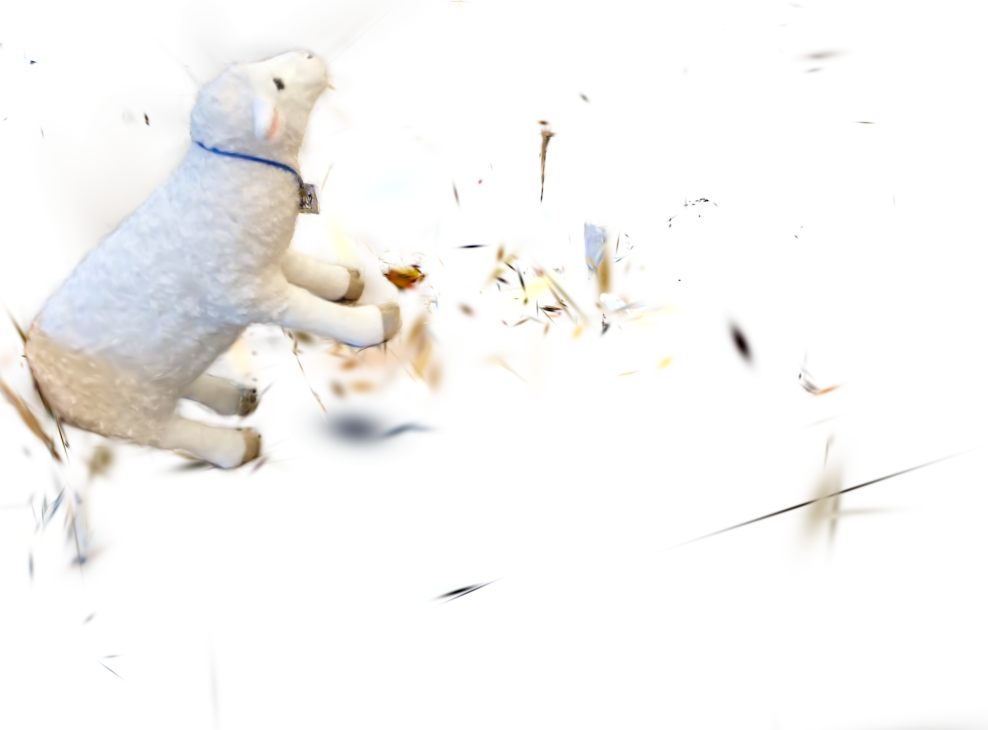} &
        \includegraphics[width=\imgw]{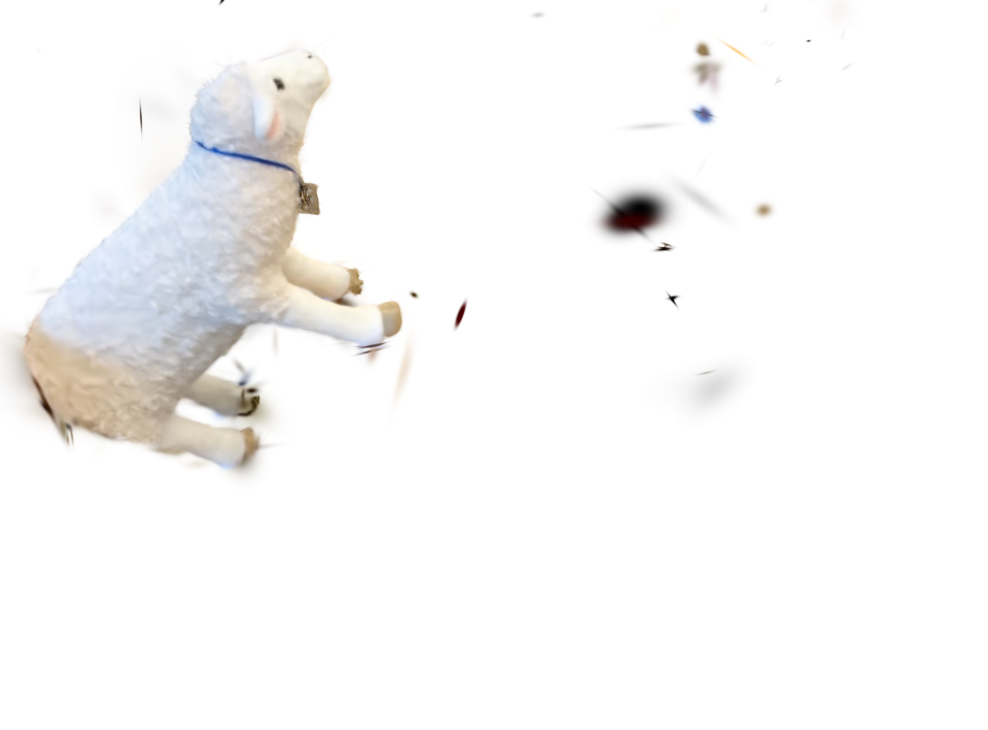} &
        \includegraphics[width=\imgw]{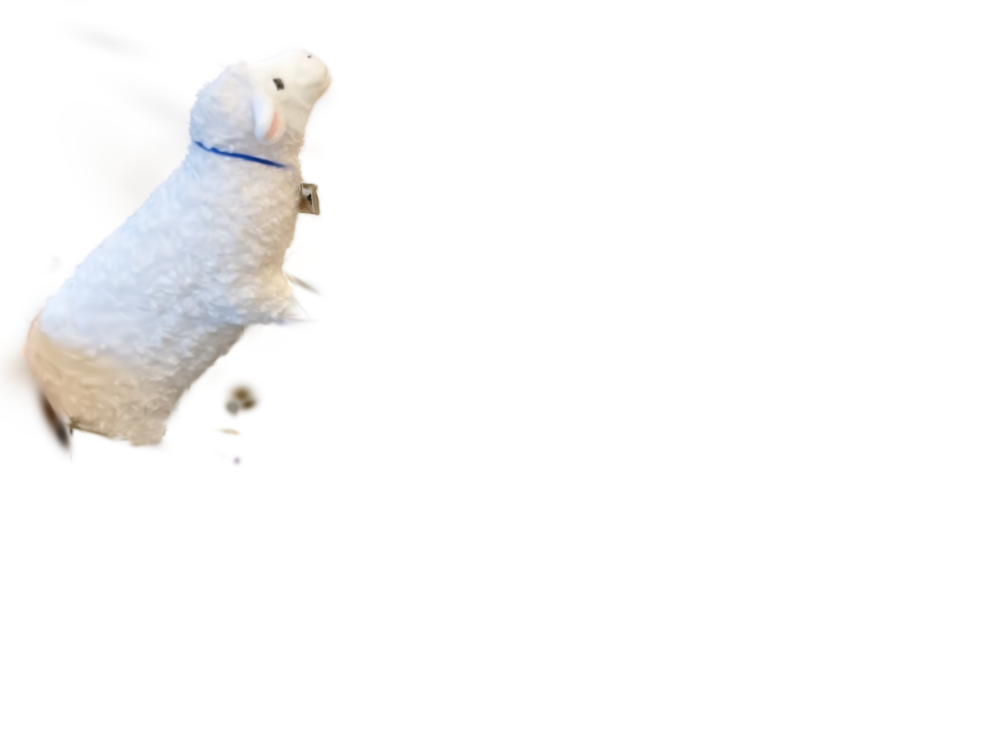} &
        \includegraphics[width=\imgw]{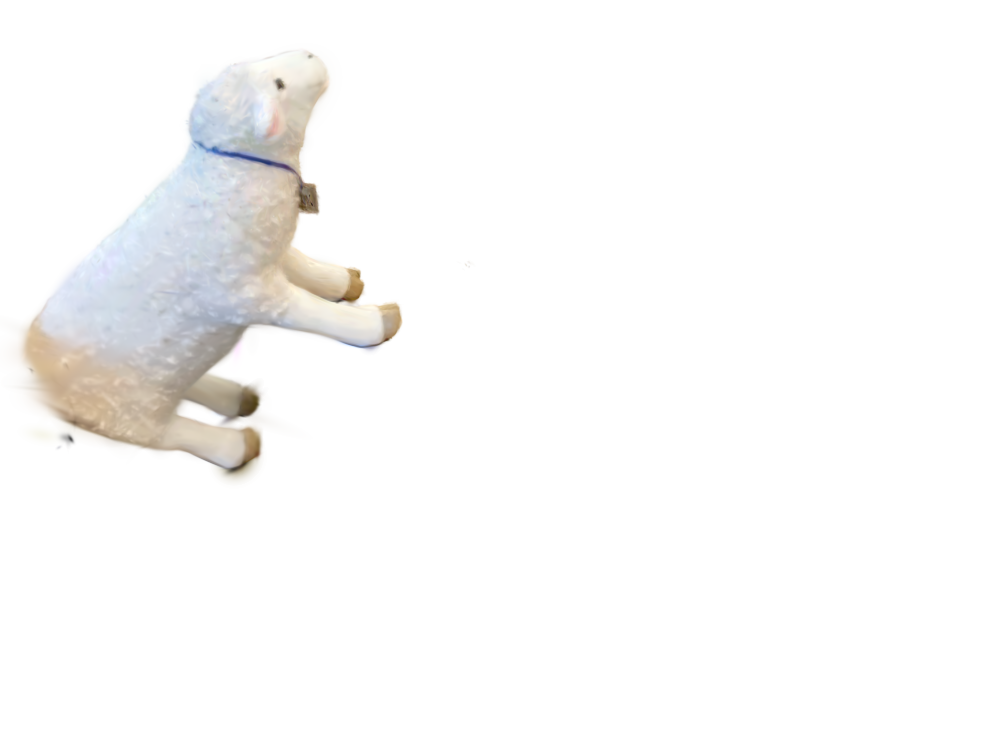} \\

    \end{tabular}
    }
    \caption{
    \textbf{Qualitative examples from the LERF dataset -- }
    We show example renders, with the selected objects marked with red arrows.
    Our method provides the best extractions, even for occluded objects and in challenging scenarios.
    }
    \label{fig:qualitative}
\end{figure*}
\begin{figure*}[t]
    \centering
    \footnotesize
    \newcommand{\imgh}{0.1\linewidth}
    \newcommand{\imghfix}{0.088\linewidth}
    \newcommand{\imghtwo}{0.2\linewidth}
    \setlength{\tabcolsep}{2pt}
    \resizebox{\linewidth}{!}{
    \begin{tabular}{cccccccc}
        & \textbf{Train Image} & \textbf{GT} & \textbf{Gaussian Grouping} & \textbf{SAGA} & \textbf{OmniSeg3D-GS} & \multicolumn{1}{c}{\textbf{Our Method}} \\
        \cmidrule(lr){2-2} \cmidrule(lr){3-3} \cmidrule(lr){4-4} \cmidrule(lr){5-5} \cmidrule(lr){6-6} \cmidrule(lr){7-7}
        \raisebox{5pt}{\rotatebox{90}{livingroom}} &
        \includegraphics[height=\imgh]{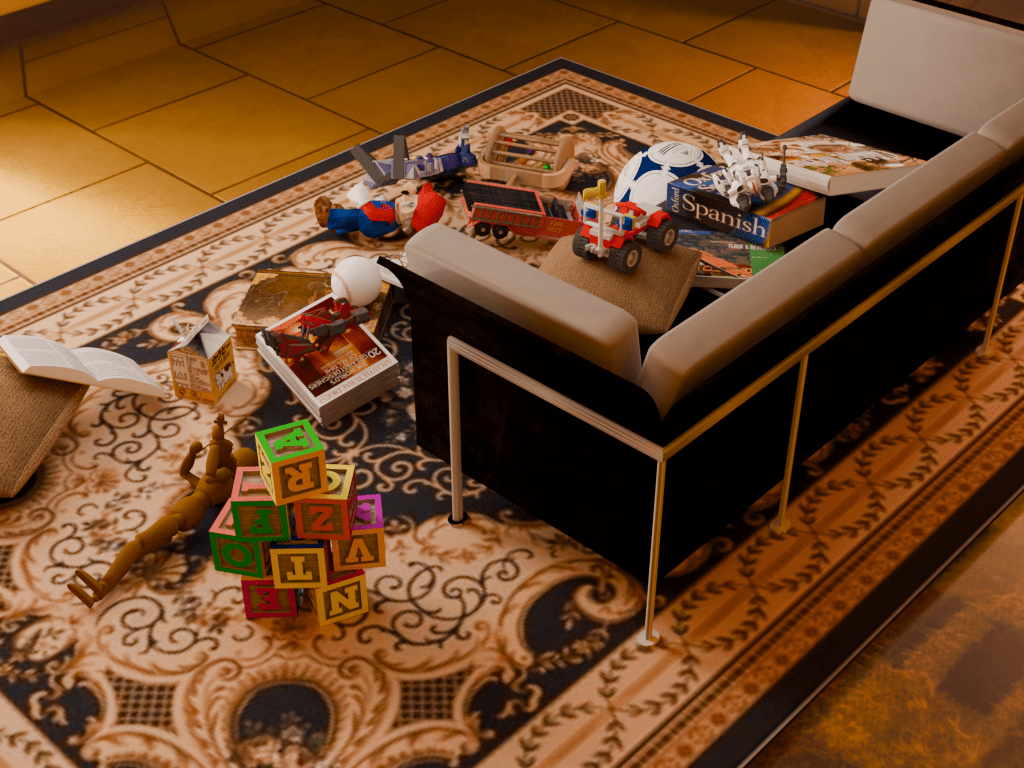} &
        \includegraphics[height=\imgh]{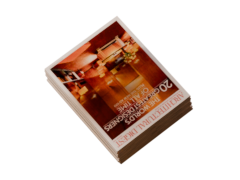} &
        \includegraphics[height=\imgh]{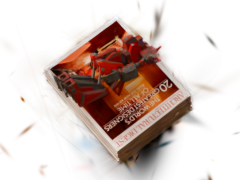} &
        \includegraphics[height=\imgh]{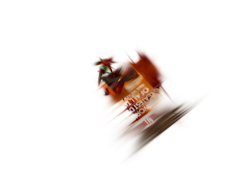} &
        \includegraphics[height=\imgh]{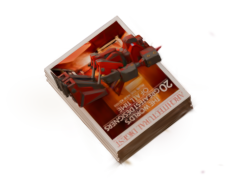} &
        \includegraphics[height=\imgh]{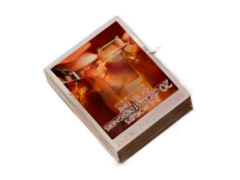}\\

        \raisebox{12pt}{\rotatebox{90}{kitchen}} &
        \includegraphics[height=\imgh]{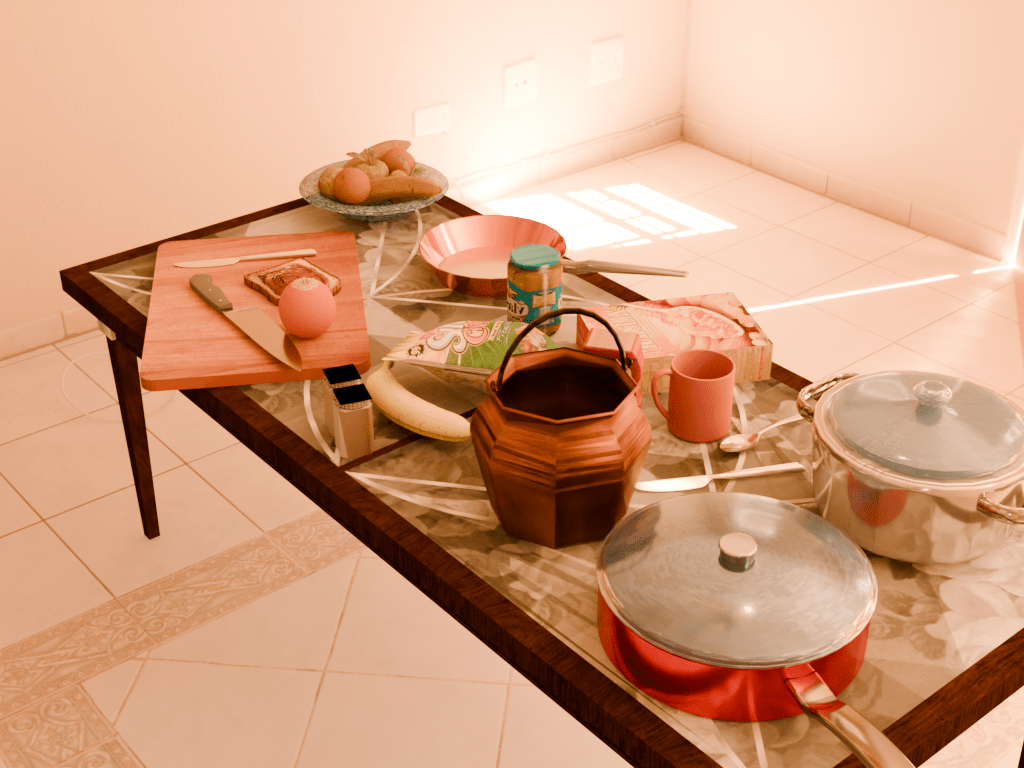} &
        \includegraphics[height=\imgh]{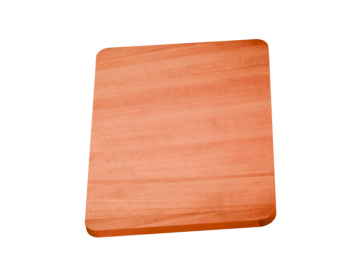} &
        \includegraphics[height=\imgh]{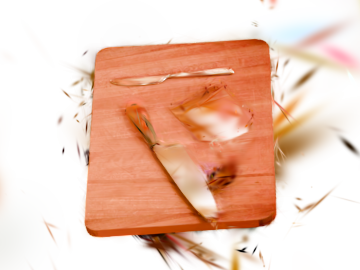} &
        \includegraphics[height=\imgh]{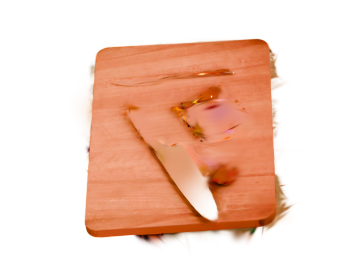} &
        \includegraphics[height=\imgh]{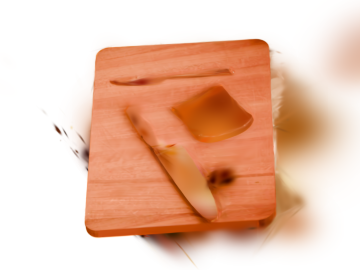} &
        \includegraphics[height=\imgh]{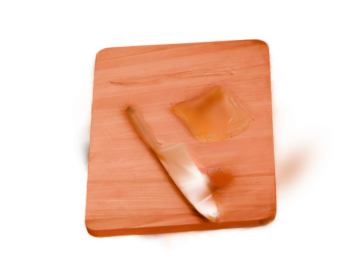} \\
        
        \raisebox{10pt}{\rotatebox{90}{bedroom}} & 
        \includegraphics[height=\imgh]{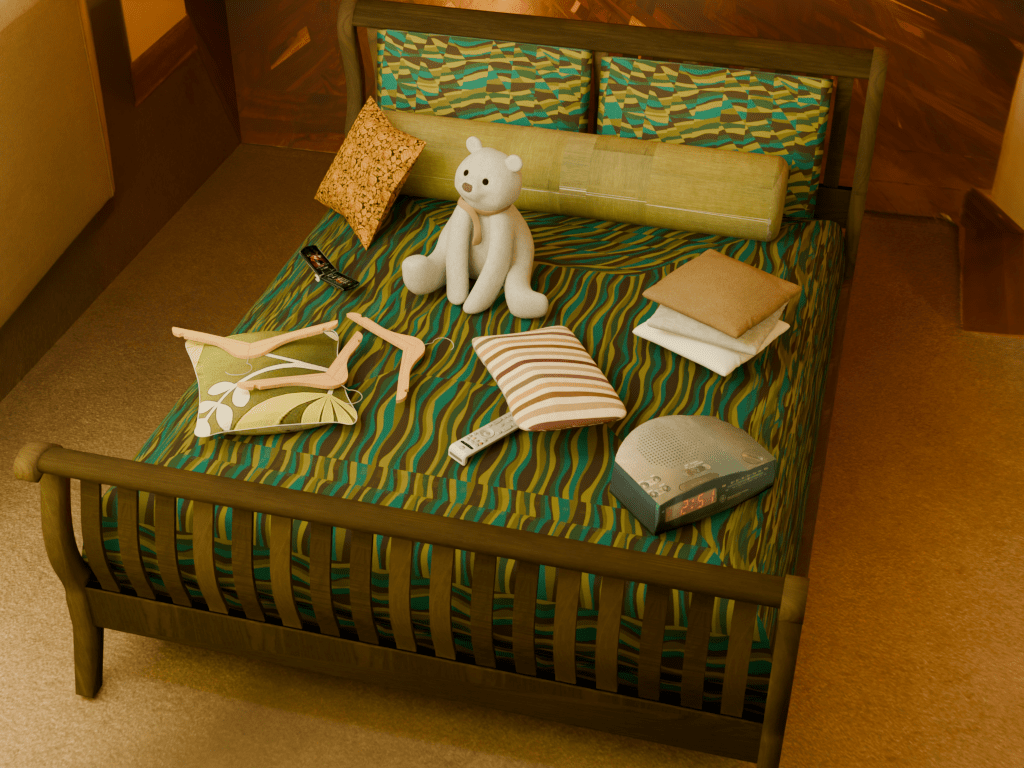} &
        \includegraphics[height=\imgh]{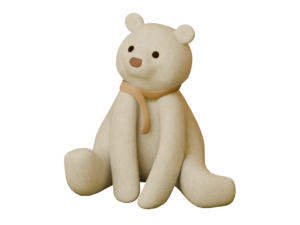} &
        \includegraphics[height=\imgh]{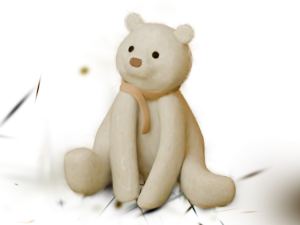} &
        \includegraphics[height=\imgh]{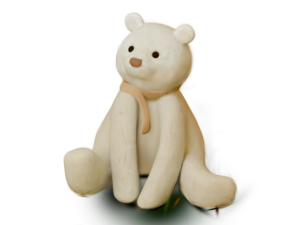} &
        \includegraphics[height=\imgh]{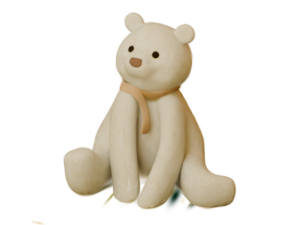} &
        \includegraphics[height=\imgh]{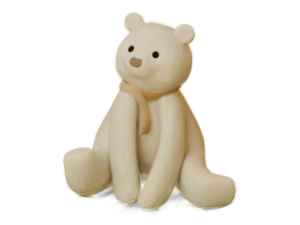} \\ 

        \raisebox{15pt}{\rotatebox{90}{picnic}} &  
        \includegraphics[height=\imgh]{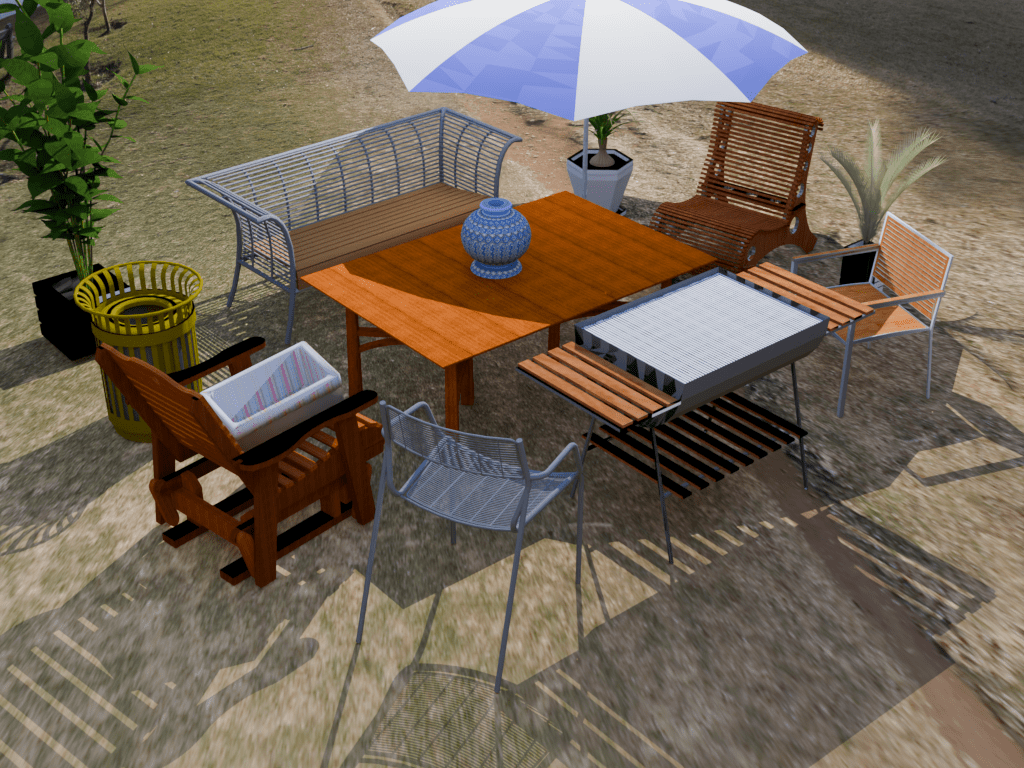} &
        \includegraphics[height=\imgh]{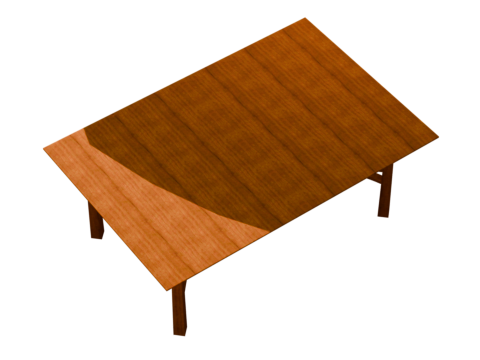} &
        \includegraphics[height=\imgh]{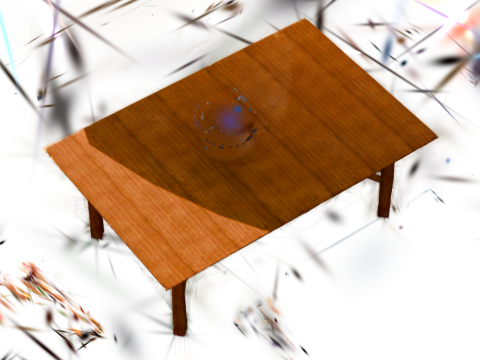} &
        \includegraphics[height=\imgh]{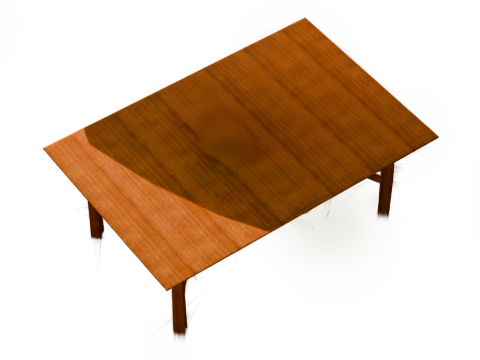} &
        \includegraphics[height=\imgh]{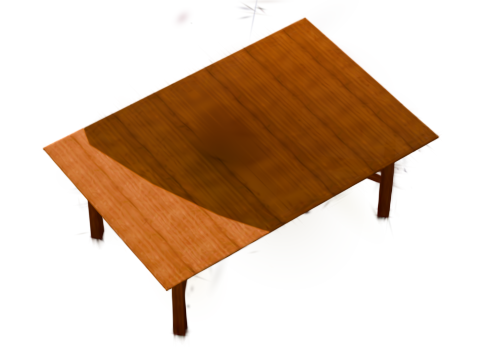} &
        \includegraphics[height=\imgh]{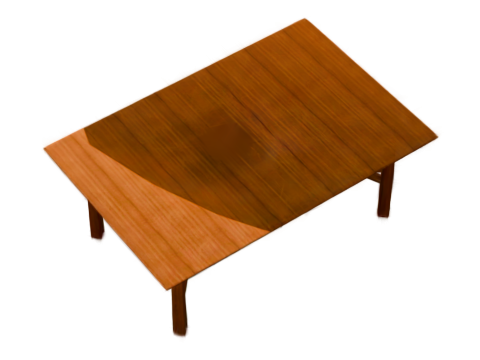} \\
        
    \end{tabular}
    }
    \caption{
    \textbf{Qualitative examples for \ourdatasetname~-- }
    We show both the training image and the object-only renders from each method including the ground-truth, and their zoom-ins.
    While all methods do moderately well when the full object is visible, as in the teddy bear object in the `bedroom' sequence, or for the table in the `picnic' sequence, in more complicated cases such as the `kitchen' or the `livingroom' sequences existing methods do not deliver best results.
    As shown in the `kitchen' sequence, our method also does not always give perfect results---we inherit the limitations of what the inpainter can do.
    As our method does not have any constraints on what inpainter can be used, we expect our results to improve as inpainters improve.
    }
    \label{fig:qualitative2}
\end{figure*}

\begin{table}[t]
    \centering
    \small
    \setlength{\tabcolsep}{1mm}
    \begin{tabular}{@{}l cccc @{}}
    \toprule    
    Method & PSNR $\uparrow$ & SSIM $\uparrow$ & LPIPS $\downarrow$ & IoU $\uparrow$ \\
    \midrule
    OmniSeg3D-GS (2024)
    & 
    \underline{34.47} & \underline{0.7837} & \underline{0.1893} & 0.5388 \\
    Gaussian Grouping (2024)
    & 
    31.17 & 0.6732 & 0.3100 & 0.3012 \\
    SAGA (2025)
    & 
    32.78 & 0.7302 & 0.2387 & \underline{0.5664} \\
    \midrule
    Ours & 
    \textbf{36.12} & \textbf{0.9164} & \textbf{0.0802} & \textbf{0.9419} \\
    \bottomrule
    \end{tabular}
    \caption{
    \textbf{Quantitative summary for \ourdatasetname~-- }
    We report quantitative summary over all objects for each scene, as well as the average.
    Our method significantly outperforms all methods.
    }    
    \label{tab:comparison}
\end{table}

\begin{table}[t]
    \centering
    \setlength{\tabcolsep}{1mm}
    \small
    \begin{tabular}{@{}l cccc @{}}
    \toprule    
    Method & PSNR $\uparrow$ & SSIM $\uparrow$ & LPIPS $\downarrow$ & IoU $\uparrow$ \\
    \midrule
        w/o-two-stage & 30.59 & 0.6538 & 0.3501 & 0.3031 \\
        w/o-inpainting & 36.02 & \textbf{0.9193} & \textbf{0.0755} & \underline{0.9418} \\
        w/o-pruning    & 35.94 & 0.9116 & 0.0874 & 0.9255 \\
        w/o-Wasserstein (Euclidean) & \underline{36.05} & 0.9144 & {0.0816} & 0.9372 \\
        Ours  & \textbf{36.12} &  \underline{0.9164} & \underline{0.0802} & \textbf{0.9419} \\
    \bottomrule
    \end{tabular} 
    \caption{
    \textbf{Quantitative ablation for \ourdatasetname --} We report ablation with various components disabled. 
    Our full method performs best.
    }    
    \label{tab:ablation}
\end{table}

\begin{figure}
    \centering
    \newcommand{\imgw}{0.32\textwidth}
    \setlength{\tabcolsep}{1pt}
    
    \begin{subfigure}{0.99\linewidth}
    \centering
    \begin{tabular}{ccc} {w/o pruning} & {w/o inpainting} &{Our Method} \\

    \includegraphics[width=\imgw]{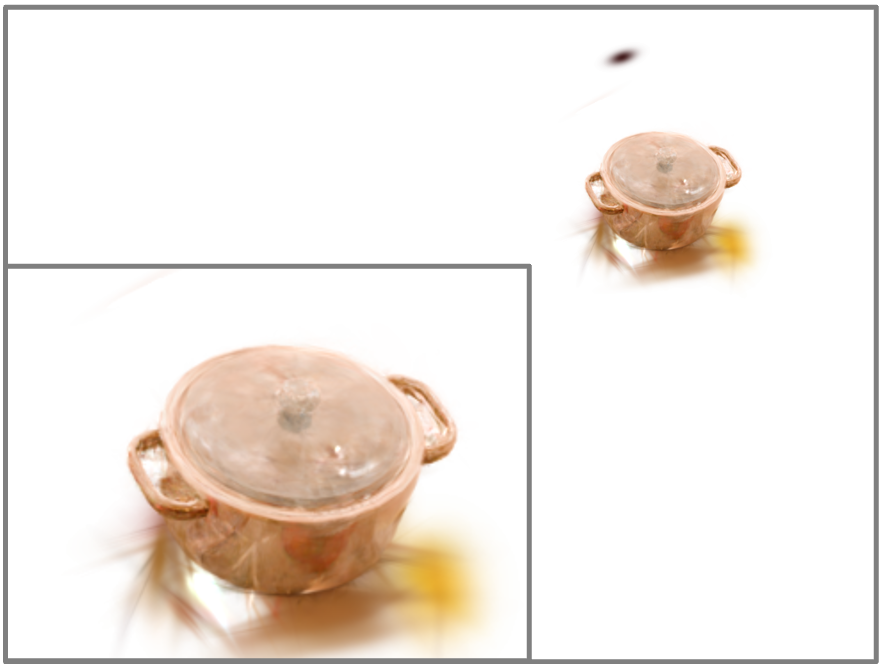} &
    \includegraphics[width=\imgw]{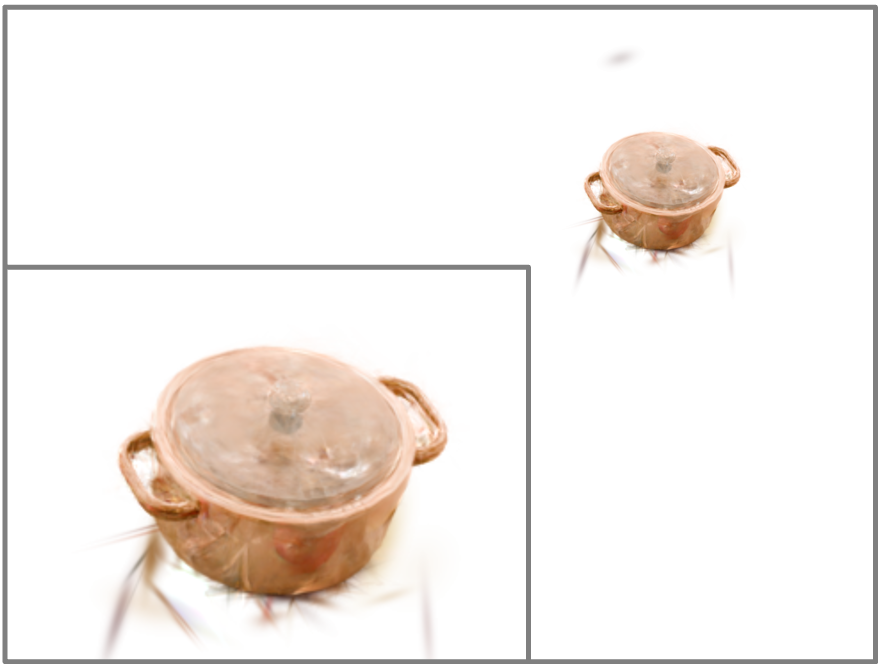} &
    \includegraphics[width=\imgw]{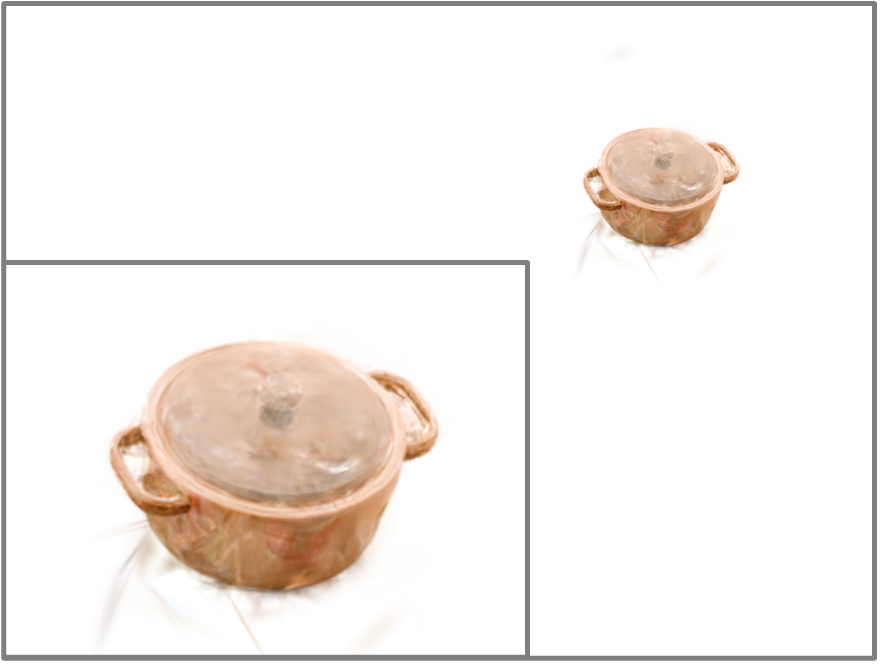} \\

    \end{tabular}
    \end{subfigure}
    
    \begin{subfigure}{0.99\linewidth}
    \centering
    \begin{tabular}{ccc} {w/o inpainting} & {\footnotesize{Non-inpaint Enhance}} & {Our Method} \\

    \includegraphics[width=\imgw]{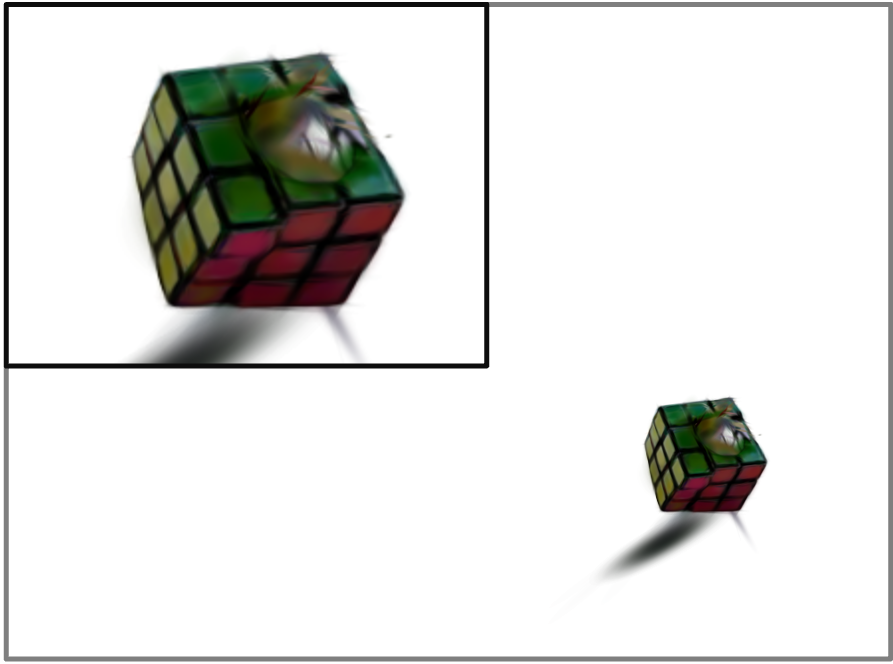} &
    \includegraphics[width=\imgw]{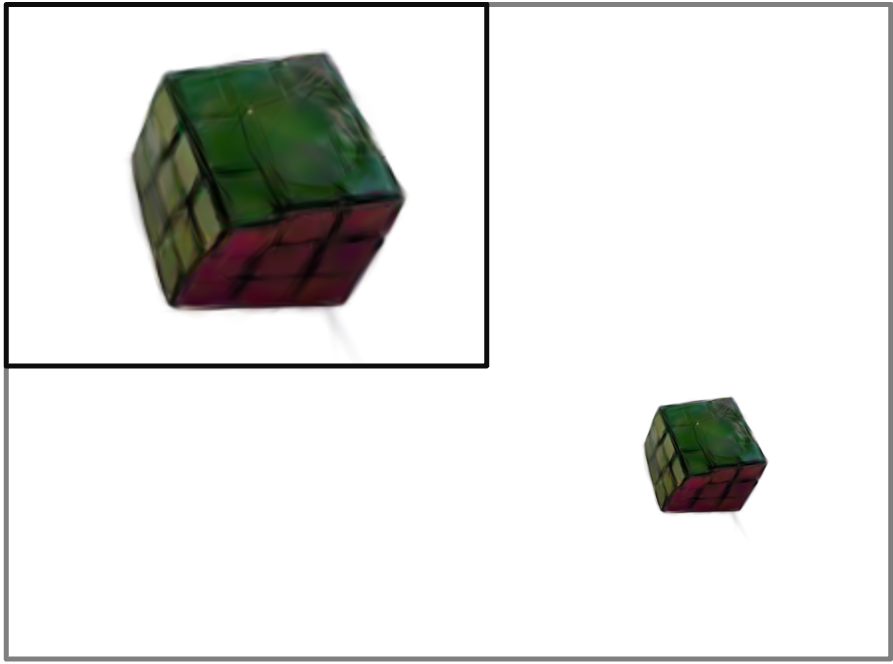} &
    \includegraphics[width=\imgw]{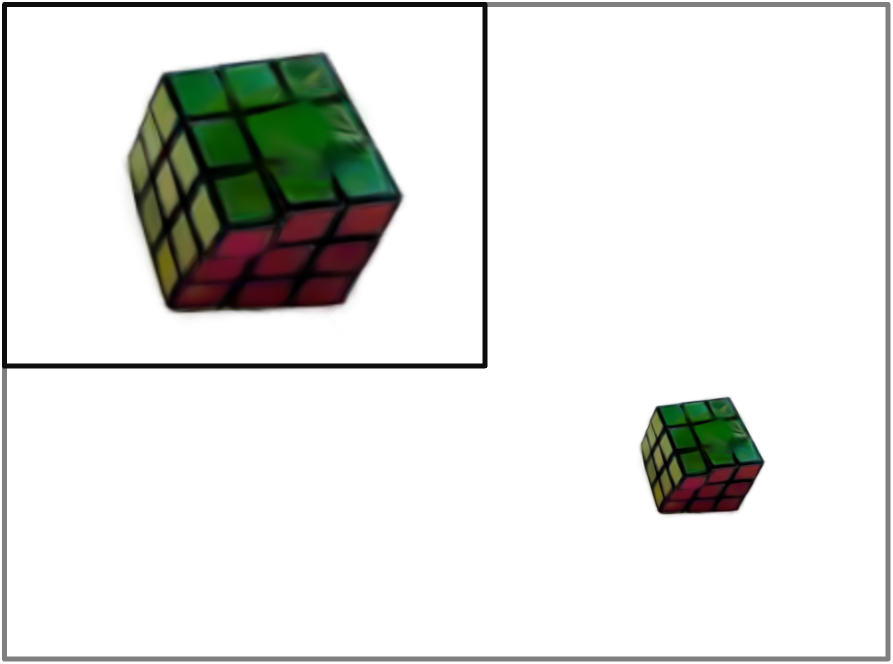} \\

    \end{tabular}
    \end{subfigure}
    \caption{
    \textbf{Ablation study -- }
    \textbf{(top row)} Numerous floaters remain without pruning. 
    Pruning alone (w/o inpainting) still leaves floaters near the pot.
    Pruning and inpainting together results in clean edges.
    \textbf{(bottom row)} Without inpainting, occluded objects exhibit holes. 
    Both non-inpainting enhancement and inpainting fill these holes, but non-inpainting enhancement also edits other correct regions, drifting away from the actual object appearance.
    }
    \label{fig:ablation}
\end{figure}

\subsection{Experimental setup}

We evaluate our method on two datasets: the LERF dataset~\cite{lerf}, a commonly used real-world dataset for extraction tasks; and on our custom synthetic dataset, which we name \ourdatasetname.

\customparagraph{LERF~\cite{lerf} dataset.}
The LERF dataset contains 14 scenes, among which we evaluate our method on three scenes as in prior work~\cite{gaussiangrouping, lifting, clickgaussian}---we use  \textit{figurines}, \textit{ramen}, and \textit{teatime}.
Being a real-world dataset, the ground-truth views for a single object are non-trivial to obtain because of occlusion.
A strightforward option is to take the masks of the object of interest on the ground-truth images and evaluate within the masks, but this would ignore the quality of reconstructions outside the masks---\eg, the floaters.
Due to this, existing works~\cite{gaussiangrouping, clickgaussian} evaluate rendered semantic features against the semantic segmentation masks, but this is not applicable to our case as we also reconstruct occluded regions.
We thus use this dataset only for qualitative evaluation, and rely on our synthetic dataset for quantitative evaluation.

\customparagraph{\ourdatasetname dataset.}
To quantitatively evaluate the quality of novel-view synthesis, even for occluded regions, we create a new synthetic dataset, \ourdatasetname.
We use the ShapeNet~\cite{shapenet} assets and render four complex scenes as shown in {fig:multiobjectblender}.
Each scene contains 5 to 20 objects, and we select 3 to 4 objects per scene for testing marked with red arrows.
Each scene has distinct characteristics, such as a large number of highly reflective objects, strong shadows from sunlight in outdoor settings, or significant occlusions where objects overlap.
To allow quantitative evaluation, we render both the full scene and the object of interest in isolation, without other objects or backgrounds.
We render 50 training images of the full scene and 50 test images for each extraction target object.

\customparagraph{Evaluation metrics}
We evaluate the quality of extraction using four standard metrics: Peak Signal-to-Noise Ratio (PSNR), Structural Similarity Index Metric (SSIM)~\cite{wang2004image}, Learned Perceptual Image Patch Similarity (LPIPS)~\cite{zhang2018unreasonable}, and Intersection over Union (IoU).
As we wish to focus on the object of interest, we only compute PSNR, SSIM, and LPIPS within a bounding box defined by 1.2$\times$ the bounding box that tightly bounds the target object.
We average the metrics over all extracted objects in each scene, then over scenes.

\customparagraph{Baselines}
We compare our method with three baselines:
\begin{itemize}
    \item \textbf{OmniSeg3D-GS~\cite{OmniSeg3D}}: 
    is an author-provided extension of OmniSeg3D~\cite{OmniSeg3D} that uses 3DGS as the 3D representation.
    It trains 3DGS with an additional feature that can be used to segment objects.
    We select a pixel on the target object and extract Gaussians whose cosine similarity of semantic feature with the target object exceeds a preset threshold.
    We carefully choose the threshold and the pixel for best results.
    \item \textbf{Gaussian Grouping~\cite{gaussiangrouping}}:  
    trains 3DGS with an additional feature, and with a classifier that takes each feature as input and outputs logits. 
    We extract Gaussians where the logit corresponding to the target is maximized.
    \item \textbf{SAGA~\cite{saga}}:
    trains 3DGS with an additional feature, similar to the former two. 
    As in the case of OmniSeg3D-GS, we again select a pixel on the target object and extract Gaussians whose cosine similarity exceeds a preset threshold.
\end{itemize}
We use the implementation provided by the authors for all baselines.
Kindly refer to the appendix for the implementation details of our method.
We will release code.

\subsection{Results}

\customparagraph{Qualitative results -- \cref{fig:qualitative,fig:qualitative2}.}
We first present qualitative results.
We show rendered images of extracted objects on the LERF dataset in \cref{fig:qualitative} and on the \ourdatasetname dataset in \cref{fig:qualitative2}.
In both datasets, our method provides improved renderings, with less floaters and holes.
Note especially the top of the cube and the lower half of the egg in \cref{fig:qualitative}, which is occluded in all views, or the book in \cref{fig:qualitative2}, where other methods struggle to reconstruct.

\customparagraph{Quantitative results -- \cref{tab:comparison}.}
We provide quantitative results for the \ourdatasetname dataset in \cref{tab:comparison}.
As reported, our method significantly outperforms all baselines.
Notably, the gap is larger when looking into SSIM and LPIPS, which reflects the quality difference shown in \cref{fig:qualitative,fig:qualitative2}.
In the case of PSNR, the gap is relatively smaller as our metric still contains fully white backgrounds to account for also floaters, which are prevalent when trying to extract objects in cluttered scenes.

\subsection{Ablation studies -- \cref{tab:ablation}, \cref{fig:ablation}.}
\label{sec:ablation}

To motivate our design choices, we perform ablation studies.
We show the importance of our statistical pruning and inpainting qualitatively in \cref{fig:ablation}.
As shown, our method without pruning results in many floaters. 
Our method without inpainting is also suboptimal, as nearby floaters remain after pruning, causing unclear object edges.
The two together provide the best results.
\cref{tab:ablation} quantifies this; the complete pipeline shows the best results for PSNR and IoU, and second-best for SSIM and LPIPS, where w/o-inpainting attains a marginally higher score due to its sensitivity to minor chromatic shifts.

We also show that using a non-inpainting diffusion model to enhance the quality of rendering also does not drastically improve results.
Specifically, instead of inpainting, we use text-to-image denoising model ``SD-XL 1.0-base''~\cite{podell2023sdxl} with the SDEdit algorithm~\cite{meng2021sdedit} to enhance the quality of renderings.
This, however, generates enhanced images that are very inconsistent from different views due to lack of preservation of the unoccluded content. 
This results in quality degradation and can even drift away from how the original object is observed in some cases.

\section{Conclusion}
We presented a novel approach for extracting objects from 3D Gaussian Splatting models. 
To do so, we have proposed a prune-and-inpaint method that applies statistical outlier removal of non-object floaters, and a scene-geometry-based occlusion reasoning method for inpainting.
We demonstrated the synergy between pruning and inpainting, both of which substantially enhance extraction performance. 
Our evaluations on both the standard LERF dataset and our new synthetic \ourdatasetname~ dataset confirm the effectiveness of our approach in producing clear object extractions that are robust to occlusions and floaters. 

\customparagraph{Limitations.}
While our method shows promising results, the quality currently is limited by the generative inpainting model. 
As research in this area is rapidly evolving, we expect this limitation to be mitigated in the future.

\bibliography{main}

\begin{thebibliography}{49}
\providecommand{\natexlab}[1]{#1}

\bibitem[{{Blender Online Community}(2024)}]{blender}
{Blender Online Community}. 2024.
\newblock {Blender - a 3D modelling and rendering package}.
\newblock \url{https://www.blender.org/}.

\bibitem[{Cen et~al.(2025)Cen, Fang, Yang, Xie, Zhang, Shen, and Tian}]{saga}
Cen, J.; Fang, J.; Yang, C.; Xie, L.; Zhang, X.; Shen, W.; and Tian, Q. 2025.
\newblock {Segment Any 3D Gaussians}.
\newblock In \emph{AAAI}.

\bibitem[{Chacko et~al.(2025)Chacko, Haeni, Khaliullin, Sun, and Lee}]{lifting}
Chacko, R.; Haeni, N.; Khaliullin, E.; Sun, L.; and Lee, D. 2025.
\newblock {Lifting by Gaussians: A Simple, Fast and Flexible Method for 3D Instance Segmentation}.
\newblock \emph{arXiv preprint}.

\bibitem[{Chang et~al.(2015)Chang, Funkhouser, Guibas, Hanrahan, Huang, Li, Savarese, Savva, Song, Su, Xiao, Yi, and Yu}]{shapenet}
Chang, A.~X.; Funkhouser, T.; Guibas, L.; Hanrahan, P.; Huang, Q.; Li, Z.; Savarese, S.; Savva, M.; Song, S.; Su, H.; Xiao, J.; Yi, L.; and Yu, F. 2015.
\newblock {ShapeNet: An Information-Rich 3D Model Repository}.
\newblock \emph{arXiv preprint}.

\bibitem[{Chen et~al.(2022)Chen, Xu, Geiger, Yu, and Su}]{chen2022tensorf}
Chen, A.; Xu, Z.; Geiger, A.; Yu, J.; and Su, H. 2022.
\newblock {TensorRF: Tensorial Radiance Fields}.
\newblock In \emph{Eur. Conf. Comput. Vis.}

\bibitem[{Chen et~al.(2024{\natexlab{a}})Chen, Chen, Zhang, Wang, Yang, Wang, Cai, Yang, Liu, and Lin}]{chen2024gaussianeditor}
Chen, Y.; Chen, Z.; Zhang, C.; Wang, F.; Yang, X.; Wang, Y.; Cai, Z.; Yang, L.; Liu, H.; and Lin, G. 2024{\natexlab{a}}.
\newblock {GaussianEditor: Swift and Controllable 3D Editing with Gaussian Splatting}.
\newblock In \emph{IEEE Conf. Comput. Vis. Pattern Recog.}

\bibitem[{Chen et~al.(2024{\natexlab{b}})Chen, Wang, Wang, and Liu}]{chen2024text}
Chen, Z.; Wang, F.; Wang, Y.; and Liu, H. 2024{\natexlab{b}}.
\newblock {Text-to-3D using Gaussian Splatting}.
\newblock In \emph{IEEE Conf. Comput. Vis. Pattern Recog.}

\bibitem[{Choi et~al.(2024)Choi, Song, Kim, Kim, and Do}]{clickgaussian}
Choi, S.; Song, H.; Kim, J.; Kim, T.; and Do, H. 2024.
\newblock {Click-Gaussian: Interactive Segmentation to Any 3D Gaussians}.
\newblock In \emph{Eur. Conf. Comput. Vis.}

\bibitem[{Fridovich-Keil et~al.(2023)Fridovich-Keil, Meanti, Warburg, Recht, and Kanazawa}]{fridovich2023k}
Fridovich-Keil, S.; Meanti, G.; Warburg, F.~R.; Recht, B.; and Kanazawa, A. 2023.
\newblock {K-planes: Explicit Radiance Fields in Space, Time, and Appearance}.
\newblock In \emph{IEEE Conf. Comput. Vis. Pattern Recog.}

\bibitem[{Fridovich-Keil et~al.(2022)Fridovich-Keil, Yu, Tancik, Chen, Recht, and Kanazawa}]{fridovich2022plenoxels}
Fridovich-Keil, S.; Yu, A.; Tancik, M.; Chen, Q.; Recht, B.; and Kanazawa, A. 2022.
\newblock {Plenoxels: Radiance Fields without Neural Networks}.
\newblock In \emph{IEEE Conf. Comput. Vis. Pattern Recog.}

\bibitem[{Gao et~al.(2024)Gao, Xu, Cao, Mildenhall, Ma, Chen, Tang, and Neumann}]{gao2024gaussianflow}
Gao, Q.; Xu, Q.; Cao, Z.; Mildenhall, B.; Ma, W.; Chen, L.; Tang, D.; and Neumann, U. 2024.
\newblock {GaussianFlow: Splatting Gaussian Dynamics for 4D Content Creation}.
\newblock \emph{arXiv preprint}.

\bibitem[{Hu et~al.(2022)Hu, Liu, Chen, Shen, and Jia}]{hu2022efficientnerf}
Hu, T.; Liu, S.; Chen, Y.; Shen, T.; and Jia, J. 2022.
\newblock {EfficientNeRF: Efficient Neural Radiance Fields}.
\newblock In \emph{IEEE Conf. Comput. Vis. Pattern Recog.}

\bibitem[{Ji et~al.(2024)Ji, Wu, Fang, Cen, Yi, Liu, Tian, and Wang}]{ji2024segment}
Ji, S.; Wu, G.; Fang, J.; Cen, J.; Yi, T.; Liu, W.; Tian, Q.; and Wang, X. 2024.
\newblock {Segment Any 4D Gaussians}.
\newblock \emph{arXiv preprint}.

\bibitem[{Kerbl et~al.(2023)Kerbl, Kopanas, Leimk{\"u}hler, and Drettakis}]{3dgs}
Kerbl, B.; Kopanas, G.; Leimk{\"u}hler, T.; and Drettakis, G. 2023.
\newblock {3D Gaussian Splatting for Real-Time Radiance Field Rendering}.
\newblock \emph{ACM Trans. Graph.}, 42.

\bibitem[{Kerr et~al.(2023)Kerr, Kim, Goldberg, Kanazawa, and Tancik}]{lerf}
Kerr, J.; Kim, C.~M.; Goldberg, K.; Kanazawa, A.; and Tancik, M. 2023.
\newblock {LERF: Language Embedded Radiance Fields}.
\newblock In \emph{Int. Conf. Comput. Vis.}

\bibitem[{Khatib and Giryes(2024)}]{khatib2024trinerflet}
Khatib, R.; and Giryes, R. 2024.
\newblock {TriNeRFLet: A Wavelet Based Triplane NeRF Representation}.
\newblock In \emph{Eur. Conf. Comput. Vis.}

\bibitem[{Kheradmand et~al.(2025)Kheradmand, Rebain, Sharma, Sun, Tseng, Isack, Kar, Tagliasacchi, and Yi}]{kheradmand20253d}
Kheradmand, S.; Rebain, D.; Sharma, G.; Sun, W.; Tseng, Y.-C.; Isack, H.; Kar, A.; Tagliasacchi, A.; and Yi, K.~M. 2025.
\newblock {3D Gaussian Splatting as Markov Chain Monte Carlo}.
\newblock In \emph{Adv. Neural Inform. Process. Syst.}

\bibitem[{Kim et~al.(2024)Kim, Wu, Kerr, Goldberg, Tancik, and Kanazawa}]{garfield}
Kim, C.~M.; Wu, M.; Kerr, J.; Goldberg, K.; Tancik, M.; and Kanazawa, A. 2024.
\newblock {GARField: Group Anything with Radiance Fields}.
\newblock In \emph{IEEE Conf. Comput. Vis. Pattern Recog.}

\bibitem[{Kirillov et~al.(2023)Kirillov, Mintun, Ravi, Mao, Rolland, Gustafson, Xiao, Whitehead, Berg, Lo et~al.}]{sam}
Kirillov, A.; Mintun, E.; Ravi, N.; Mao, H.; Rolland, C.; Gustafson, L.; Xiao, T.; Whitehead, S.; Berg, A.~C.; Lo, W.-Y.; et~al. 2023.
\newblock {Segment Anything}.
\newblock In \emph{Int. Conf. Comput. Vis.}

\bibitem[{Li et~al.(2022)Li, Weinberger, Belongie, Koltun, and Ranftl}]{lseg}
Li, B.; Weinberger, K.~Q.; Belongie, S.; Koltun, V.; and Ranftl, R. 2022.
\newblock {Language-driven Semantic Segmentation}.
\newblock In \emph{Int. Conf. Learn. Represent.}

\bibitem[{Li et~al.(2024)Li, Chen, Li, and Xu}]{li2024spacetime}
Li, Z.; Chen, Z.; Li, Z.; and Xu, Y. 2024.
\newblock {Spacetime Gaussian Feature Splatting for Real-time Dynamic View Synthesis}.
\newblock In \emph{IEEE Conf. Comput. Vis. Pattern Recog.}

\bibitem[{Luiten et~al.(2024)Luiten, Kopanas, Leibe, and Ramanan}]{luiten2024dynamic}
Luiten, J.; Kopanas, G.; Leibe, B.; and Ramanan, D. 2024.
\newblock {Dynamic 3D Gaussians: Tracking by Persistent Dynamic View Synthesis}.
\newblock In \emph{Int. Conf. 3D Vis.}

\bibitem[{Meng et~al.(2021)Meng, He, Song, Song, Wu, Zhu, and Ermon}]{meng2021sdedit}
Meng, C.; He, Y.; Song, Y.; Song, J.; Wu, J.; Zhu, J.-Y.; and Ermon, S. 2021.
\newblock {SDEdit: Guided Image Synthesis and Editing with Stochastic Differential Equations}.
\newblock \emph{arXiv preprint}.

\bibitem[{Mildenhall et~al.(2020)Mildenhall, Srinivasan, Tancik, Barron, Ramamoorthi, and Ng}]{nerf}
Mildenhall, B.; Srinivasan, P.~P.; Tancik, M.; Barron, J.~T.; Ramamoorthi, R.; and Ng, R. 2020.
\newblock {NeRF: Representing Scenes as Neural Radiance Fields for View Synthesis}.
\newblock In \emph{Eur. Conf. Comput. Vis.}

\bibitem[{Podell et~al.(2023)Podell, English, Lacey, Blattmann, Dockhorn, M{\"u}ller, Penna, and Rombach}]{podell2023sdxl}
Podell, D.; English, Z.; Lacey, K.; Blattmann, A.; Dockhorn, T.; M{\"u}ller, J.; Penna, J.; and Rombach, R. 2023.
\newblock {SDXL: Improving Latent Diffusion Models for High-Resolution Image Synthesis}.
\newblock \emph{arXiv preprint}.

\bibitem[{Ravi et~al.(2024)Ravi, Gabeur, Hu, Hu, Ryali, Ma, Khedr, R{\"a}dle, Rolland, Gustafson et~al.}]{ravi2024sam}
Ravi, N.; Gabeur, V.; Hu, Y.-T.; Hu, R.; Ryali, C.; Ma, T.; Khedr, H.; R{\"a}dle, R.; Rolland, C.; Gustafson, L.; et~al. 2024.
\newblock {SAM 2: Segment Anything in Images and Videos}.
\newblock \emph{arXiv preprint}.

\bibitem[{Ren et~al.(2023)Ren, Pan, Tang, Zhang, Cao, Zeng, and Liu}]{ren2023dreamgaussian4d}
Ren, J.; Pan, L.; Tang, J.; Zhang, C.; Cao, A.; Zeng, G.; and Liu, Z. 2023.
\newblock {DreamGaussian4D: Generative 4D Gaussian Splatting}.
\newblock \emph{arXiv preprint}.

\bibitem[{Rusu and Cousins(2011)}]{sor}
Rusu, R.~B.; and Cousins, S. 2011.
\newblock 3d is here: Point cloud library (pcl).
\newblock In \emph{2011 IEEE international conference on robotics and automation}, 1--4. IEEE.

\bibitem[{Shen, Yang, and Wang(2024)}]{flashsplat}
Shen, Q.; Yang, X.; and Wang, X. 2024.
\newblock {FlashSplat: 2D to 3D Gaussian Splatting Segmentation Solved Optimally}.
\newblock In \emph{Eur. Conf. Comput. Vis.}

\bibitem[{Silva et~al.(2024)Silva, Dahaghin, Toso, and Del~Bue}]{contrastivegaussian}
Silva, M.~C.; Dahaghin, M.; Toso, M.; and Del~Bue, A. 2024.
\newblock {Contrastive Gaussian Clustering: Weakly Supervised 3D Scene Segmentation}.
\newblock \emph{arXiv preprint}.

\bibitem[{Sun, Sun, and Chen(2022)}]{sun2022direct}
Sun, C.; Sun, M.; and Chen, H.-T. 2022.
\newblock {Direct Voxel Grid Optimization: Super-fast Convergence for Radiance Fields Reconstruction}.
\newblock In \emph{IEEE Conf. Comput. Vis. Pattern Recog.}

\bibitem[{Tang et~al.(2023)Tang, Ren, Zhou, Liu, and Zeng}]{tang2023dreamgaussian}
Tang, J.; Ren, J.; Zhou, H.; Liu, Z.; and Zeng, G. 2023.
\newblock {DreamGaussian: Generative Gaussian Splatting for Efficient 3D Content Creation}.
\newblock \emph{arXiv preprint}.

\bibitem[{{The Diffusers team}(2023)}]{diffusers-inpainting}
{The Diffusers team}. 2023.
\newblock {SD-XL Inpainting 0.1}.

\bibitem[{Turki et~al.(2024)Turki, Agrawal, Bul{\`o}, Porzi, Kontschieder, Ramanan, Zollh{\"o}fer, and Richardt}]{turki2024hybridnerf}
Turki, H.; Agrawal, V.; Bul{\`o}, S.~R.; Porzi, L.; Kontschieder, P.; Ramanan, D.; Zollh{\"o}fer, M.; and Richardt, C. 2024.
\newblock {HybridNeRF: Efficient Neural Rendering via Adaptive Volumetric Surfaces}.
\newblock In \emph{IEEE Conf. Comput. Vis. Pattern Recog.}

\bibitem[{Wang et~al.(2004)Wang, Bovik, Sheikh, and Simoncelli}]{wang2004image}
Wang, Z.; Bovik, A.~C.; Sheikh, H.~R.; and Simoncelli, E.~P. 2004.
\newblock {Image Quality Assessment: from Error Visibility to Structural Similarity}.
\newblock \emph{IEEE Trans. Image Process.}

\bibitem[{Weber et~al.(2024)Weber, Holynski, Jampani, Saxena, Snavely, Kar, and Kanazawa}]{weber2024nerfiller}
Weber, E.; Holynski, A.; Jampani, V.; Saxena, S.; Snavely, N.; Kar, A.; and Kanazawa, A. 2024.
\newblock {Nerfiller: Completing Scenes via Generative 3D Inpainting}.
\newblock In \emph{IEEE Conf. Comput. Vis. Pattern Recog.}

\bibitem[{Wu et~al.(2024{\natexlab{a}})Wu, Yi, Fang, Xie, Zhang, Wei, Liu, Tian, and Wang}]{wu20244d}
Wu, G.; Yi, T.; Fang, J.; Xie, L.; Zhang, X.; Wei, W.; Liu, W.; Tian, Q.; and Wang, X. 2024{\natexlab{a}}.
\newblock {4D Gaussian Splatting for Real-time Dynamic Scene Rendering}.
\newblock In \emph{IEEE Conf. Comput. Vis. Pattern Recog.}

\bibitem[{Wu et~al.(2024{\natexlab{b}})Wu, Mildenhall, Henzler, Park, Gao, Watson, Srinivasan, Verbin, Barron, Poole, and Holynski}]{reconfusion}
Wu, R.; Mildenhall, B.; Henzler, P.; Park, K.; Gao, R.; Watson, D.; Srinivasan, P.~P.; Verbin, D.; Barron, J.~T.; Poole, B.; and Holynski, A. 2024{\natexlab{b}}.
\newblock {ReconFusion: 3D Reconstruction with Diffusion Priors}.
\newblock In \emph{IEEE Conf. Comput. Vis. Pattern Recog.}

\bibitem[{Wu et~al.(2024{\natexlab{c}})Wu, Yu, Jiang, Cao, Wang, and Bai}]{wu2024sc4d}
Wu, Z.; Yu, C.; Jiang, Y.; Cao, C.; Wang, F.; and Bai, X. 2024{\natexlab{c}}.
\newblock {SC4D: Sparse-controlled Video-to-4D Generation and Motion Transfer}.
\newblock In \emph{Eur. Conf. Comput. Vis.} Springer.

\bibitem[{Yang et~al.(2024)Yang, Gao, Zhou, Jiao, Zhang, and Jin}]{yang2024deformable}
Yang, Z.; Gao, X.; Zhou, W.; Jiao, S.; Zhang, Y.; and Jin, X. 2024.
\newblock {Deformable 3D Gaussians for High-fidelity Monocular Dynamic Scene Reconstruction}.
\newblock In \emph{IEEE Conf. Comput. Vis. Pattern Recog.}

\bibitem[{Yang et~al.(2023)Yang, Yang, Pan, and Zhang}]{yang2023real}
Yang, Z.; Yang, H.; Pan, Z.; and Zhang, L. 2023.
\newblock {Real-time Photorealistic Dynamic Scene Representation and Rendering with 4D Gaussian Splatting}.
\newblock \emph{arXiv preprint}.

\bibitem[{Ye et~al.(2024)Ye, Danelljan, Yu, and Ke}]{gaussiangrouping}
Ye, M.; Danelljan, M.; Yu, F.; and Ke, L. 2024.
\newblock {Gaussian Grouping: Segment and Edit Anything in 3D Scenes}.
\newblock In \emph{Eur. Conf. Comput. Vis.}

\bibitem[{Yi et~al.(2024)Yi, Fang, Wang, Wu, Xie, Zhang, Liu, Tian, and Wang}]{yi2024gaussiandreamer}
Yi, T.; Fang, J.; Wang, J.; Wu, G.; Xie, L.; Zhang, X.; Liu, W.; Tian, Q.; and Wang, X. 2024.
\newblock {GaussianDreamer: Fast Generation from Text to 3D Gaussians by Bridging 2D and 3D Diffusion Models}.
\newblock In \emph{IEEE Conf. Comput. Vis. Pattern Recog.}

\bibitem[{Ying et~al.(2024)Ying, Yin, Zhang, Wang, Yu, Huang, and Fang}]{OmniSeg3D}
Ying, H.; Yin, Y.; Zhang, J.; Wang, F.; Yu, T.; Huang, R.; and Fang, L. 2024.
\newblock {OmniSeg3D: Omniversal 3D Segmentation via Hierarchical Contrastive Learning}.
\newblock In \emph{IEEE Conf. Comput. Vis. Pattern Recog.}

\bibitem[{Yu et~al.(2021)Yu, Li, Tancik, Li, Ng, and Kanazawa}]{yu2021plenoctrees}
Yu, A.; Li, R.; Tancik, M.; Li, H.; Ng, R.; and Kanazawa, A. 2021.
\newblock {PlenOctrees for Real-time Rendering of Neural Radiance Fields}.
\newblock In \emph{Int. Conf. Comput. Vis.}

\bibitem[{Zeng et~al.(2024)Zeng, Jiang, Zhu, Lu, Lin, Zhu, Hu, Cao, and Yao}]{zeng2024stag4d}
Zeng, Y.; Jiang, Y.; Zhu, S.; Lu, Y.; Lin, Y.; Zhu, H.; Hu, W.; Cao, X.; and Yao, Y. 2024.
\newblock {STAG4D: Spatial-temporal Anchored Generative 4D Gaussians}.
\newblock In \emph{Eur. Conf. Comput. Vis.} Springer.

\bibitem[{Zhang et~al.(2018)Zhang, Isola, Efros, Shechtman, and Wang}]{zhang2018unreasonable}
Zhang, R.; Isola, P.; Efros, A.~A.; Shechtman, E.; and Wang, O. 2018.
\newblock {The Unreasonable Effectiveness of Deep Features as a Perceptual Metric}.
\newblock In \emph{IEEE Conf. Comput. Vis. Pattern Recog.}

\bibitem[{Zhang, Chen, and Cui(2025)}]{zhang2025efficient}
Zhang, Y.; Chen, G.; and Cui, S. 2025.
\newblock {Efficient Large-scale Scene Representation with a Hybrid of High-resolution Grid and Plane Features}.
\newblock \emph{Pattern Recognition}.

\bibitem[{Zhou et~al.(2024)Zhou, Chang, Jiang, Fan, Zhu, Xu, Chari, You, Wang, and Kadambi}]{feature3dgs}
Zhou, S.; Chang, H.; Jiang, S.; Fan, Z.; Zhu, Z.; Xu, D.; Chari, P.; You, S.; Wang, Z.; and Kadambi, A. 2024.
\newblock {Feature 3DGS: Supercharging 3D Gaussian Splatting to Enable Distilled Feature Fields}.
\newblock In \emph{IEEE Conf. Comput. Vis. Pattern Recog.}

\end{thebibliography}


\end{document}